\documentclass[journal]{IEEEtran}

\usepackage{amsmath,graphicx,multirow,subfigure}
\usepackage{rotating,comment}
\usepackage{hyperref}
\usepackage{cite}
\usepackage{makecell}
\usepackage{amssymb}
\usepackage{multirow}
\usepackage{algpseudocode,algorithm}
\usepackage{color}

\ifCLASSINFOpdf

\else

\fi

\usepackage{graphicx}

\usepackage{fancyhdr}
\begin{document}
%
\title{Improving fingerprint presentation attack detection by an approach integrated into the personal verification stage}
%
%
%

\author{Marco~Micheletto\textsuperscript{\textsection},~\IEEEmembership{Member,~IEEE,} Giulia~Orrù\textsuperscript{\textsection},~\IEEEmembership{Member,~IEEE,} \\Luca~Ghiani, and~Gian Luca~Marcialis,~\IEEEmembership{Senior~Member,~IEEE}
\IEEEcompsocitemizethanks{\IEEEcompsocthanksitem Marco Micheletto, Giulia Orrù and Gian Luca Marcialis are with  University of Cagliari,
09123 Cagliari, Italy (e-mail: marco.micheletto@unica.it; giulia.orru@unica.it; marcialis@unica.it).
\IEEEcompsocthanksitem Luca Ghiani is with University of Sassari, 07100 Sassari, Italy (e-mail: lghiani@uniss.it). \protect\\}
\thanks{This work has been submitted to the IEEE for possible publication. Copyright may be transferred without notice, after which this version may no longer be accessible.}}

%
%

\markboth{Journal of \LaTeX\ Class Files,~Vol.~14, No.~8, August~2015}%
{Shell \MakeLowercase{\textit{et al.}}: Bare Demo of IEEEtran.cls for IEEE Journals}
%



\maketitle

\begin{abstract}
Presentation Attack Detection (PAD) systems are usually designed independently of the fingerprint verification system. 
While this can be acceptable for use cases where specific user templates are not predetermined, it represents a missed opportunity to enhance security in scenarios where integrating PAD with the fingerprint verification system could significantly leverage users' templates, which are the real target of a potential presentation attack. This does not mean that a PAD should be specifically designed for such users; that would imply the availability of many enrolled users' PAI and, consequently, complexity, time, and cost increase. On the contrary, we propose to equip a basic PAD, designed according to the state of the art, with an innovative add-on module called the Closeness Binary Code (CC) module.
The term ``closeness'' refers to a peculiar property of the \textit{bona fide}-related features: in an Euclidean feature space, genuine fingerprints tend to cluster in a specific pattern. First, samples from the same finger are close to each other, then samples from other fingers of the same user and finally, samples from fingers of other users. This property is statistically verified in our previous publication, and further confirmed in this paper. It is independent of the user population and the feature set class, which can be handcrafted or deep network-based (embeddings). Therefore, the add-on can be designed without the need for the targeted user samples; moreover, it exploits her/his samples' ``closeness'' property during the verification stage. Extensive experiments on benchmark datasets and state-of-the-art PAD methods confirm the benefits of the proposed add-on, which can be easily coupled with the main PAD module integrated into the fingerprint verification system.
\end{abstract}


%
\IEEEpeerreviewmaketitle

\section{Introduction}
\IEEEPARstart{F}{ingerprint} presentation attack detection (PAD) \cite{marcel2019handbook} aims to identify when an individual tries to deceive an Automated Fingerprint Identification System (AFIS) by presenting a fingerprint replica made up of artificial materials and technically named ``presentation attack instrument'' (PAI) instead of the actual \textit{bona fide} finger \cite{iso}. This defensive mechanism is crucial for ensuring the integrity and reliability of biometric authentication systems.
In recent years, various innovative PAD approaches have been proposed, achieving impressive performance on benchmarking datasets \cite{micheletto2023review}. 
However, the prevailing design philosophy often treats PAD as an independent module from the AFIS, assuming that presentation attacks are uniform across different users. Even more, they disregard the variations between the bona-fide access samples coming from different users. This approach might be practical for general applications where the primary goal is to detect the authenticity of fingerprints without linking them to specific individual identities, ignoring the integration of PAD within the AFIS framework. In circumstances where such knowledge is available and relevant, where the availability of bona-fide samples of the subject is expected, such as in personal identity verification systems (e.g. on personal devices) or in small and medium-scale verification systems, it constitutes a significant oversight. 
\\
In such scenarios, it is possible to design a PAD system tailored to the user's unique biometric data. While this idea has demonstrated efficacy in facial recognition domains, enhancing security through user-specific adaptations \cite{chingovska2015on}, its application within fingerprint biometrics, to the best of our knowledge, has yet to be thoroughly explored. This may be attributed to the demanding data requirements of a user-specific PAD system. Developing an effective system would necessitate a comprehensive collection of diverse PAI materials and attack methodologies specific to each user: a task that poses significant challenges in terms of privacy, cost, and logistical feasibility \cite{marcel2019handbook}. 

\begin{figure}
    \centering
\includegraphics[width=0.7\linewidth]{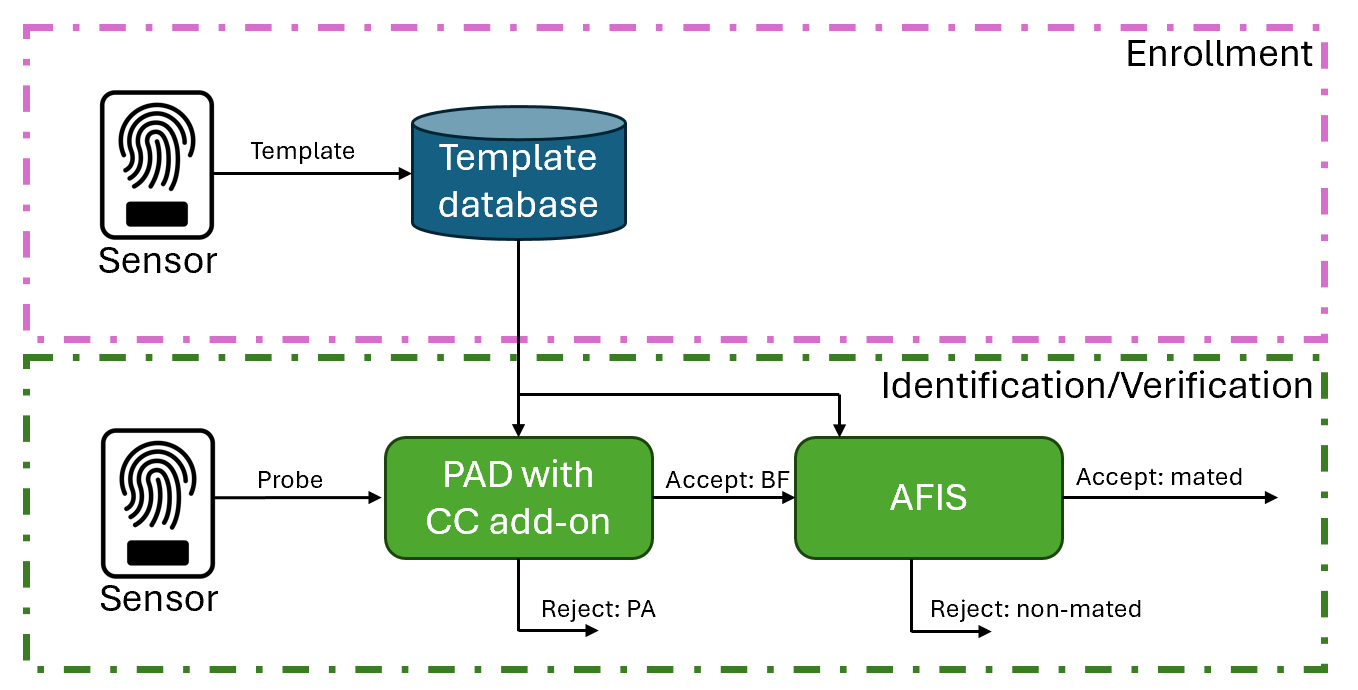}
    \caption{Workflow illustrating the application of the Closeness Binary Code (CC) module in verification scenarios. During enrollment, user fingerprint templates are stored in the database. In the verification stage, the PAD system, enhanced by the CC add-on, processes input samples to classify them as bona fide (BF) or presentation attacks (PA) exploiting template database information. Accepted samples are further compared in the AFIS to verify identity, enabling robust security in small-to-medium-scale setups or personal devices like smartphones.}
    \label{fig:usecase}
\end{figure}

To circumvent these challenges, we propose a viable alternative. We can consider a standard PAD and conceive an additional module, namely, an add-on. This is integrated between the PAD and the AFIS, and exploits the AFIS's enrolled users' information, without re-train or re-fine the PAD in any aspect. This way is much simpler than applying transfer learning or domain transformation algorithms, where a significant set of novel and user-specific data must be available. As shown in Figure \ref{fig:usecase}, the system is designed for environments where live samples are readily available, such as smartphones or small-to-medium-scale setups.

Accordingly, we propose a methodology whose basis is the findings of our previous paper, Ref. \cite{ghiani2017fingerprint}. In particular, we show that it is possible to exploit the \textit{bona fide} data acquired during the enrolment phase (template gallery) as auxiliary information for presentation attack detection. Ref. \cite{ghiani2016user} has shown that the handcrafted features adopted for presentation attack detection also contain specific information about the related subject. Ghiani \textit{et al.} \cite{ghiani2016user} associated these properties with the skin characteristics of all the subject's fingers and the unique characteristics of the individual fingerprint, also referred to as ``person-specific'' and ``finger-specific'', respectively. The observed consequence was that \textit{bona fide} features of the same user are geometrically close; nevertheless, they are distant from those of other users, which are partially overlapped with presentation attacks. 
\\
Our paper extends these findings to deep-learning features, the so-called ``embeddings''. Then, we design a module that can be incorporated into any PAD model, called ``Closeness Binary Code'' (CC). This module aims to empower any presentation attack detector ability when integrated into the AFIS. 

In addition, this work:
\begin{itemize}
    \item extends the application of CC to white-box and black-box handcrafted and deep-learning PAD models to highlight its effectiveness on modern PADs. In \cite{ghiani2016user}, CC was only applied on white-box PADs consisting of handcrafted features and an SVM classifier (that work was published in 2016). With the term white-box models, we refer to models whose implementation details are available;
    \item demonstrates the utility of the CC method on novel scenarios, with realistic PAI generation techniques \cite{casula2023towards};
    \item simulates the integration of PADs in an AFIS to evaluate the improvement introduced by the CC method.
\end{itemize}
The paper is organized as follows. Section \ref{ref:relatwork} overviews the current literature on PAD. Section \ref{ref:userspec} describes the proposed algorithm based on the ``closeness binary code'' (CC). Section \ref{ref:exper} explores the potential of the proposed approach by experiments on white-box and black-box state-of-the-art (SOTA) PADs and their integration with a comparator.  Section \ref{ref:concl} concludes the paper.

\section{Related works}
\label{ref:relatwork}


In the field of fingerprint PAD, there has been continuous research and development aimed at addressing the vulnerability of fingerprint sensors to artificial replicas. The threat of fingerprint PAs has been recognized since 1998 \cite{willis1998six}. In response to this threat, hardware and software countermeasures were proposed in subsequent years \cite{schuckers2002spoofing}.

PAD systems have focused on extracting and utilizing anatomical, physiological, textural and other relevant features to assess whether the presentation to the sensor is \textit{bona fide} or not. The evolution of feature extraction and AI image classification in PAD modules has transitioned from hand-crafted approaches to the deep learning era. Hand-crafted extractors, such as the Scale Invariant Feature Transformation (SIFS) \cite{gottschlich2014fingerprint}, the     Binarized Statistical Image Features
(BSIF) \cite{bsifghiani}, the Local Binary Pattern (LBP) \cite{lbpfing}, and Local Phase Quantization
(LPQ) \cite{lpq}, are based on convolving the image with linear filters and binarizing the filter responses \cite{gragnaniello2013fingerprint}. These embeddings are commonly input to shallow classifiers like Support Vector Machines (SVM) \cite{cortes1995support} but can also be utilized in complex neural networks. On the other hand, deep learning techniques, mainly based on Convolutional Neural Networks (CNN), can automatically learn discriminative features, offering high accuracy albeit requiring substantial training data and computational resources \cite{grosz2020fingerprint}.

In recent years, there has been a growing trend of combining hand-crafted and deep learning-based techniques to overcome limitations associated with modern PAD systems, such as their difficulty in generalizing across different sensors and the substantial computational resources they demand. One example of this combination is the Fingerprint Spoof Buster system \cite{chugh2018fingerprint}, which employs locally centred and aligned patches utilizing fingerprint minutiae to train a MobileNet-v1 model.

The ongoing generalization problem is mainly due to the constant improvement of PAI crafting techniques. 
In fact, both hand-crafted and deep-learning solutions have been shown to suffer a significant increase in error rates when applied to PAs obtained with unknown materials \cite{sharma2022intelligent} or techniques \cite{casula2023towards}, a phenomenon known as cross-material and cross-method generalization.
Although the researchers have achieved significant advancements, including dataset diversity, transfer learning \cite{rani2021fingerprint}, adversarial learning \cite{galli2023}, data augmentation \cite{ametefe2023fingerprint}, ensemble methods \cite{sharma2022an}, and domain adaptation \cite{gonzalez2023fisher}, the generalization issue can still be considered open.

The common element across these studies is extracting information relevant to detecting presentation attacks without considering the identity class. However, in an effort to address the persistent issue of generalization, recent findings have suggested that the efficacy of PAD systems could potentially be augmented by incorporating specific biometric traits of individual users. This observation is particularly corroborated by research within the facial recognition domain, which suggests that the features critical for identifying presentation attacks are inherently linked to user-specific attributes \cite{fatemifar2019combining, fatemifar2021client, chingovska2015on}. 

For instance, Chingovska \textit{et al.} \cite{chingovska2015on} investigated the extent of user-specific information within features and its influence on system performance. Notably, their method significantly outperforms client-independent approaches. Yet, a limitation of their method is the necessity to have access to all PA samples during the enrolment phase, a condition difficult to achieve in real-world applications. Addressing this concern, Yang \textit{et al.} \cite{yang2015person} introduced a PAD strategy based on a unique classifier for each subject in order to mitigate the perturbations arising from subject variability. They further employed a subject domain adaptation technique to create virtual features, thus facilitating the training of individualized classifiers.
 
In the fingerprint domain, a similar effect has been observed, where the characteristics of the individual users notably influenced the PAD performance \cite{rattani2012analysis}. In \cite{ghiani2016user} we called this a ``finger-specific effect'', attributing it to the unique characteristics inherent to each fingerprint. Our subsequent work \cite{ghiani2017fingerprint} corroborated these findings, showing a substantial drop in error rates when a PAD system incorporated multiple acquisitions of different fingerprints of the same person during the training phase. While these studies have laid the groundwork, the full potential of leveraging user-specific biometric traits to enhance system performance in fingerprint PAD has yet to be fully realized and integrated into widespread practice.

On these bases, the next section introduces the Closeness Binary Code, a module that aims to exploit user-specific biometric characteristics to increase the efficiency and accuracy of fingerprint PAD systems. 

\section{The Closeness binary Code}
\label{ref:userspec} 
\begin{figure}
    \centering
\includegraphics[width=0.23\textwidth]{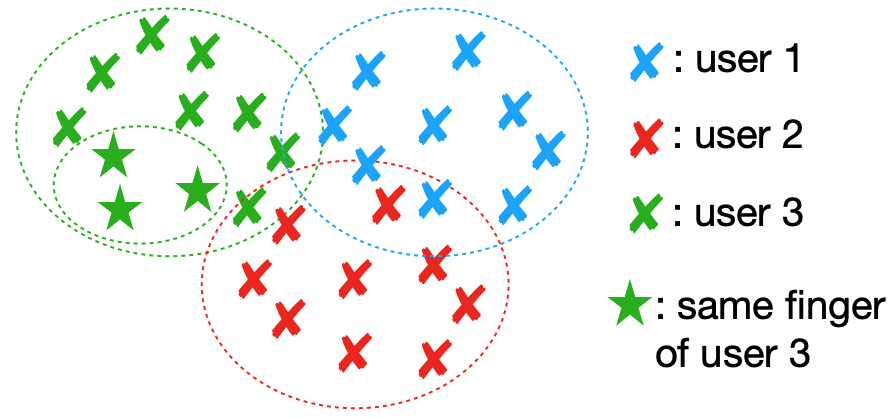}
    \caption{\textit{Bona fide} embeddings from the same user tend to be close in feature space. Knowledge of user-specific information can be used to more accurately classify new user samples into bona-fide or presentation attack, assuming it is supposed to be located within the user cluster in the former case.}
    \label{fig:geomdistr}
\end{figure}

\subsection{Motivations}
\label{sec:motivations}
The concept of user-specific variations influencing the performance of biometric systems is a fundamental principle within personal recognition. This notion, thoroughly documented in the seminal work introducing Doddington's zoo \cite{doddington1998sheep}, highlights how verification scores are affected by the individual's identity. According to this study, users are categorized into four groups based on their impact on verification system error rates. While this original study focused on speaker recognition, the principle broadly applies across biometric modalities and scenarios involving presentation attacks \cite{rattani2012analysis, fatemifar2021client}
The underlying idea is that the variability in the biometric traits among individuals could contribute to how presentation attacks manifest. For instance, the texture, color, or reflectivity of a person's skin might influence how a fake fingerprint is detected. 
This was further confirmed in the LivDet 2017 competition \cite{orru2019analysis}, where the presence of the same users, both in the train and in the test set, increased the PAD accuracy of the system.
Nevertheless, it is important to note that this boost in performance was partly due to the inclusion of users' artificial replicas in the training set. As reported in \cite{ghiani2016user}, it is not realistic to obtain a sufficiently representative and varied quantity of PAIs from the user's fingerprints in real-world applications.

This methodology aligns with the ``generic user'' standard approach where PAIs are created by specialists and are acquired and inserted, together with \textit{bona fide} samples, in the dataset used for the PAD system training. However, they do not correspond to the user population where the system must be operating, in order to avoid data biasing \cite{torralba2011unbiased}. 
For this reason, in this paper, we do not use PAIs directly coming from the user population. Instead, we expect to benefit from \textit{bona fide} samples normally collected as templates in an integrated fingerprint verification system.

In particular, we assume that the expressive power of \textit{bona fide} embeddings of the same user are geometrically close to each other and distant from that of other users. A conceptual illustration is presented in Figure \ref{fig:geomdistr}; however, a more detailed validation of this hypothesis will be provided in Section \ref{sec:hieran}.

Knowing the distribution of a user's \textit{bona fide} templates, therefore, allows the PAD to categorize a known user's sample more easily: we believe this could be even more true if we consider the same finger, which could have a high probability of being geometrically close to corresponding acquisitions. In fact, the same finger acquisitions can differ in the skin (oily, dirty, or dry) and acquisition conditions (pressure, position on the sensor), but they always share an essential part of information exploited by the comparator and compacted in the embeddings.
Based on this evidence, we hypothesize that a generic embedding is a function of three components, i.e., the generic liveness component, which we will later refer to as ``generic users'', the ``person-specific'' component, and the ``finger-specific' component. Separating these three pieces of information allows us to obtain a PAD classification approach based on a ``closeness binary code'' (CC). The CC approach describes each input image with a code indicating if the closest sample among the templates is related to a \textit{bona fide} or a presentation attack and if it represents the same finger or user. 

\subsection{The proposed add-on}
This section outlines our proposed add-on, developed under the hypothesis that the feature space is Euclidean. This assumption is crucial as it underpins our evaluation of distances and relationships within the analyzed feature spaces. At the core of our method is the Closeness binary Code (CC), represented by the triplet $\{b_f, b_p, b_o\}$,  that indicates the geometric proximity of the input sample to different types of \textit{bona fide} presentations of a validation set. The instance of these three bits is the Closeness Code associated with the input sample. Each sample is represented as a feature vector, and the selected distance agrees with the hypothesis over the related feature space (L1, L2, etc.).
In this paper, the CC module relies on a pre-trained PAD model to extract feature vectors from images. By hypothesizing they come from an L2 space, the L2 (Euclidean) distance is applied. Hence, the bits of the CC are defined as follows: 
\begin{itemize}
    \item \textit{same finger closeness} ($b_f$): This bit is set to 1 if the closest sample in the Euclidean space is a \textit{bona fide} presentation of the same finger that is being analyzed. Otherwise, it is set to 0.
    \item \textit{same person closeness} ($b_p$): This bit is set to 1 if the nearest sample in the Euclidean space is a \textit{bona fide} presentation of the same individual. If not, the bit is 0.
    \item \textit{generic user closeness} ($b_o$): This bit is set to 1 if the nearest sample in Euclidean space represents a \textit{bona fide} presentation of any other user, referred to as a ``generic user'', which is different from the identity being claimed. Otherwise, the bit is 0.
\end{itemize}
The CC computation involves an initial Look-Up Table (LUT) estimation phase, as delineated in Algorithm \ref{Alg::Training}. This phase is conducted before the actual test of the system and consequently without prior knowledge of the user population's characteristics.

For the LUT estimation, it is necessary to have a pre-trained PAD (white-box or black-box), which outputs the embedding extracted from the input image, in addition to the probability of being a \textit{bona fide} presentation. This pre-trained PAD is the one to which the CC module is added. 
It is essential to highlight that, in our design, we treat the PAD system as non-accessible, meaning we cannot modify or retrain it using its original dataset. This situation often mirrors the real-world conditions where the PAD system is acquired as a commercial product and integrated into an existing framework.
Given this constraint, an additional validation set $V$, encompassing both \textit{bona fide} and presentation attack samples, is necessary for constructing the LUT. 
For each validation sample, the PAD extracts the feature vector and computes the probability of the image being \textit{bona fide} presentations. For each feature vector $x$ derived from $V$, we compute the binary code $B(x) = \{b_f, b_p, b_o\}$ corresponding to one of eight ``closeness configurations''. The LUT is then tasked with mapping these binary codes to a class label $D(B(x))\in \{BF,PA\}$, based on the relative frequency of \textit{bona fide} and presentation attack samples within each closeness code configuration, indicated as $P(BF|b)$ and $P(PA|b)$, for each $b \in \{0 0 0,...,1 1 1\}$.
\begin{algorithm}[th]
\caption{LUT estimation starting from a validation set $V=\cup_{f \in F} \cup_{u \in U} X_f^u$ where $U=\{u_1, u_2, ..., u_n\}$ is the set of given training users, $F=\{thumb_l, index_l, ..., pinkie_r\}$ the set of possible reference fingers of the user (ten) and $X_f^u=\{x_1, x_2, ..., x_m\}$ the set of embeddings extracted 
from the same finger $f\in F$ of the client $u\in U$.
Let $L_f^u=\{l_1, l_2,..., l_m\}$ be a set of class labels $l_i\in \{BF, PA\}$, for each $f\in F$ of the client $u\in U$.
Let $N(BF|b)$ be the absolute frequency of BF samples  associated to the closeness binary code $b \in \{000, 001, 010, ..., 111\}$ and $A(b)$ the overall correct classification rate achieved by the pre-trained model on the training samples corresponding to the code $b$.}
\label{Alg::Training}
\begin{algorithmic}
\For {$b \in \{ 000, 001, 010, ..., 111\}$}
\State $N(BF|b) \leftarrow 0$
\State $N(PA|b) \leftarrow 0$
\EndFor
\For {$u \in U$}
\For {$f \in F$}
\For {$x \in X_f^u$}
\State Let $x_i$ be the closest pattern to $x$ from $V \setminus \{x\}$.
\State Let $x_j$ be the closest pattern to $x$ from $V \setminus X_f^u$. 
\State Let $x_k$ be the closest pattern to $x$ from $V\setminus \cup_{f\in F}  X_{f}^u$.
\State $b_f \leftarrow (l_i==BF) \textbf{and} (x_i \in X_f^u)$
\State $b_p \leftarrow (l_j==BF) \textbf{and} (x_j \in \cup_{f'\in F\setminus \{ f\}} X_{f'}^u)$
\State $B(x) \leftarrow \{ b_f, b_p, l_k==BF\}$
\If {$l_x==BF$}
\State $N(BF|B(x)) \leftarrow N(BF|B(x)) + 1$
\Else
\State $N(PA|B(x)) \leftarrow N(PA|B(x)) + 1$
\EndIf
\EndFor   
\EndFor 
\EndFor
\For {$b \in \{001, 010, ..., 110\}$}
\State $P(BF|b) = \frac{N(BF|b)}{N(BF|b)+N(PA|b)}$
\State $P(PA|b) = 1-P(BF|b)$
\If {$P(BF|b)<A(b)$ \textbf{and} $P(PA|b)<A(b)$}
\State $D(b) \leftarrow pre-trainedModelDecision$
\Else
\If {$P(BF|b) > P(PA|b)$}
\State $D(b) \leftarrow BF$
\Else
\State $D(b) \leftarrow PA$
\EndIf
\EndIf
\EndFor
\State $D(0 0 0) \leftarrow PA$
\State $D(1 1 1) \leftarrow BF$

\end{algorithmic}
\end{algorithm}
Based on the functioning assumptions of the method described in Section \ref{sec:motivations}, we expect that the configuration 0 0 0 always corresponds to the PA class, while the configuration 1 1 1 corresponds to the BF class. For this reason, we force the 0 0 0 configuration on the PA class and the 1 1 1 configuration on the BF class regardless of the result of the LUT estimation phase. For the other configurations, the maximum number of validation samples of the most frequent class is evaluated. If this number is higher than the number of samples correctly classified by the pre-trained model (A(b)), this class is automatically associated with the configuration. Otherwise, the pre-trained model will classify the samples that fall into the configuration.

\begin{figure}
    \centering    \includegraphics[width=1\linewidth]{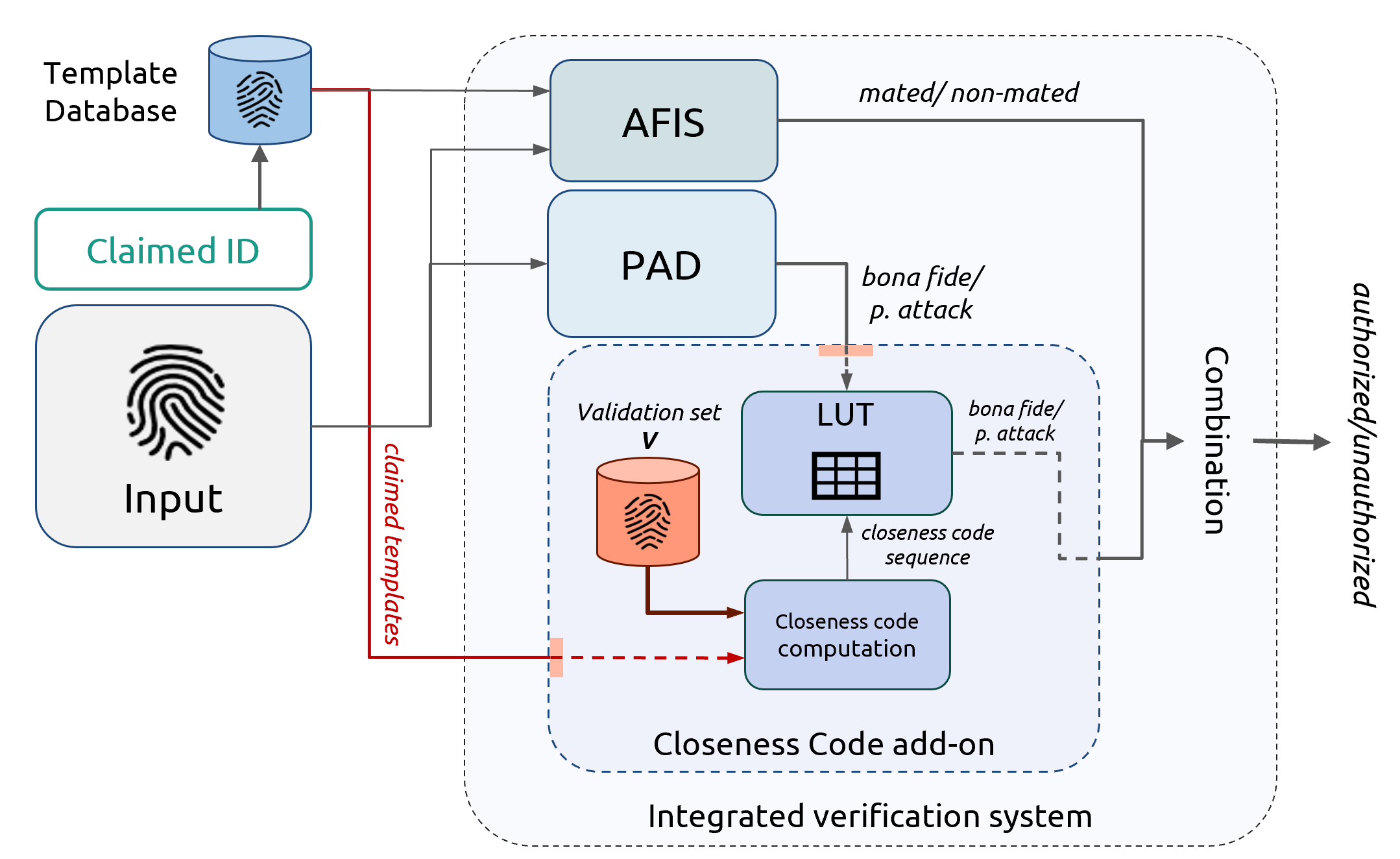}
    \caption{Overview of an AFIS with the Closeness binary Code add-on. The system combines the claimed identity templates with the validation set $V$ to compute the closeness sequence. The corresponding outcome from the Look-Up Table (LUT) is then integrated with the AFIS decision to derive the final authorization status.}
    \label{fig:cc_scheme}
\end{figure}

The classification phase of the system, outlined in Algorithm \ref{Alg::Testing} and visualized in Figure \ref{fig:cc_scheme}, constitutes the verification stage. This stage integrates user-specific information by inserting users' \textit{bona fide} templates within the validation set $V$. The CC module then computes the distances between the input sample and the expanded validation set $V$, generating the novel closeness code. Finally, the LUT is referenced to classify the sample as \textit{bona fide} or presentation attack. An illustrative example is provided to elucidate the operational dynamics of the system.
Table \ref{table:lutex} provides explicit numerical outcomes from our LUT estimation, demonstrating how the algorithm behaves for different configurations of the closeness binary code ($b_f$, $b_p$, $b_o$).
Firstly, we note the cases where our algorithm's predictions align with the expected outcomes; when the closeness binary code is 0 0 0, the classification is correctly forced to PA, reflecting the high probability (99.14\%) of the input sample's distance from any \textit{bona fide} presentation. 
In contrast, for configuration 1 1 1, the sample is close to \textit{bona fide} presentations, resulting in a correct forced categorization into the BF class.

However, the adaptability of our algorithm is particularly evident in scenarios with mixed configurations. For instance, consider the case 0 0 1. In this situation, the sample is close to a generic user's \textit{bona fide} presentation ($b_o=1$), resulting in a $P(BF|b)$ of 84.54\%. Since it is lower than A(b) (92.75\%), as outlined in our decision process, the final decision is left to the pre-trained PAD decision.

Alternatively, let us consider the case of 0 1 0. In this scenario, the sample is only close to a \textit{bona fide} presentation from the same person ($b_p=1$). Nonetheless, this proximity results in a $P(BF|b)$ of 85.00\%, exceeding A(b) of 65.00\%, thereby leading to the sample's classification as BF. Similar considerations can be made for the remaining $b$ configurations.

Subsequently, in the classification phase, the CC of the input sample is obtained by calculating the distance between it and the set obtained from the union of the validation set and the user's template gallery. The class to be associated with this sample is decided by consulting the LUT.
The CC module can be used as a supplementary module of any PAD, allowing the exploitation of the template gallery's information to correct the PAD's classification.
\begin{table}[]
\caption{Example of Look-Up Table (LUT) estimation starting from a generic pretrained PAD and a generic validation set $V$. Each row represents a different CC configuration $b$ and demonstrates how the algorithm decides between \textit{bona fide} (BF) and presentation attack (PA) based on calculated probabilities ($P(BF|b)$ and $P(PA|b)$) and the accuracy of the baseline PAD ($A(b)$).}
\centering
\label{table:lutex}
\begin{tabular}{|ccccc|}
\hline
\multicolumn{5}{|c|}{\textbf{LUT Estimation}} \\ \hline
\multicolumn{1}{|c|}{\textit{b}} & \multicolumn{1}{c|}{\textit{$P(BF|b)$}} & \multicolumn{1}{c|}{\textit{$P(PA|b)$}} & \multicolumn{1}{c|}{\textit{A(b)}} & \textit{D(b)}                                   \\ \hline
\multicolumn{1}{|c|}{0 0 0}               & \multicolumn{1}{c|}{0.86}                        & \multicolumn{1}{c|}{99.14}                       & \multicolumn{1}{c|}{80.95}                  & forced to PA                                             \\ \hline
\multicolumn{1}{|c|}{0 0 1}               & \multicolumn{1}{c|}{84.54}                       & \multicolumn{1}{c|}{15.46}                       & \multicolumn{1}{c|}{92.75}                  & \begin{tabular}[c]{@{}c@{}}$P(BF|b)<A(b)$  and\\ $P(PA|b)<A(b)$\\ →pretrained PAD decision\end{tabular} \\ \hline
\multicolumn{1}{|c|}{0 1 0}               & \multicolumn{1}{c|}{85.00}                       & \multicolumn{1}{c|}{15.00}                       & \multicolumn{1}{c|}{65.00}                  & $P(BF|b)>A(b)$→BF                                        \\ \hline
\multicolumn{1}{|c|}{0 1 1}               & \multicolumn{1}{c|}{100.00}                      & \multicolumn{1}{c|}{0.00}                        & \multicolumn{1}{c|}{94.12}                  & $P(BF|b)>A(b)$→BF                                         \\ \hline
\multicolumn{1}{|c|}{1 0 0}               & \multicolumn{1}{c|}{98.00}                       & \multicolumn{1}{c|}{2.00}                        & \multicolumn{1}{c|}{90.00}                  & $P(BF|b)>A(b)$→BF                                        \\ \hline
\multicolumn{1}{|c|}{1 0 1}               & \multicolumn{1}{c|}{99.83}                       & \multicolumn{1}{c|}{0.17}                        & \multicolumn{1}{c|}{97.62}                  & $P(BF|b)>A(b)$→BF                                        \\ \hline
\multicolumn{1}{|c|}{1 1 0}               & \multicolumn{1}{c|}{100.00}                      & \multicolumn{1}{c|}{0.00}                        & \multicolumn{1}{c|}{91.30}                  & $P(BF|b)>A(b)$→BF                                       \\ \hline
\multicolumn{1}{|c|}{1 1 1}               & \multicolumn{1}{c|}{100.00}                      & \multicolumn{1}{c|}{0.00}                        & \multicolumn{1}{c|}{96.10}                  & forced to BF                                             \\ \hline
\end{tabular}
\end{table}

\begin{algorithm}[H]
\caption{Classification stage of a unknown sample $x$ of claimed identity $w \in W=\{w_1, w_2, ..., w_n\}$ ($W \cap U = \emptyset$) and finger $f \in F$, submitted to the fingerprint verification system. Let $V=\cup_{f \in F} \cup_{u \in U} X_f^u$ be the validation set of the LUT estimation (see Alg. \ref{Alg::Training}). Let $X_f^w=\{x_1, x_2, ..., x_n\}$ be a set of embeddings extracted from the same finger $f\in F$ of the client $w\in W$. These \textit{bona fide} samples correspond to the templates stored in the fingerprint verification system. Let $D(b)$ be the decision associated to the closeness binary code $b \in \{000, 001, 010, ..., 111\}$, estimated in the training stage.}
\label{Alg::Testing}
\small
\begin{algorithmic}
\State Let $x_i$ be the closest pattern to $x$ from $X_f^w \cup V$.
\State Let $x_j$ be the closest pattern to $x$ from $(\cup_{f' \in F\setminus \{f\}} X_{f'}^w) \cup V$. 
\State Let $x_k$ be the closest pattern to $x$ from $V$.
\State $b_f \leftarrow (l_i==BF) \textbf{and} (x_i \in X_f^w)$
\State $b_p \leftarrow (l_j==BF) \textbf{and} (x_j \in \cup_{f'\in F\setminus \{ f\}} X_{f'}^w)$
\State $B(x) \leftarrow \{b_f, b_p, l_k==BF\}$
\State $d \leftarrow D(B(x))$
\If {$d==pre-trainedModelDecision$}
\State Leave the decision to the pre-trained model.
\EndIf
\end{algorithmic}
\end{algorithm}

\section{Experimental evaluation}
\label{ref:exper}
\subsection{Experimental protocol}
To evaluate the effectiveness of adding the CC module to a presentation attack detection system, we divided the experiments into two experimental protocols:
\begin{itemize}
    \item \textit{white-box protocol}: we used fully controllable PADs in this protocol. In particular, an handcrafted PAD consisting of an SVM classifier with features extracted with a the BSIF textural method \cite{bsifghiani}, and a simple CNN-based PAD were used. The CNN architecture consists of five convolutional and pooling layers followed by a flattened layer and a dense layer for classification. From the flattened layer, we extract a 2048-element embedding. The CNN model is trained to operate on cropped images with respect to the center of the fingerprint region of interest.
    \item \textit{black-box protocol}: we employed algorithms from the international LivDet 2021 competition in this protocol, specifically MEGVII\footnote{https://www.youtube.com/watch?v=WrlR1XFdyXU\&t=43s} and PADUnk21 \cite{gonzalez2021local}. MEGVII, the competition winner, is a deep learning-based model renowned for its top-tier performance, making it a SOTA tool in PAD. On the other hand, PADUnk21, while not achieving high scores, is valued for its handcrafted nature, offering a contrasting approach to MEGVII's deep learning technique. It is worth noting that these models while providing cutting-edge PAD capabilities, are not entirely controllable as they are pre-trained: for each image, they just return a probability of being \textit{bona fide} and the corresponding embedding.
\end{itemize}

Since the black-box models are pre-trained on the LivDet 2021 training set, we used the same set to train the white-box models. Respecting the specifications of the LivDet competition, where the PADs were presented, we chose an acceptance threshold of $thr=0.5$ for the evaluations.

\begin{figure}
    \centering
    \includegraphics[width=1\linewidth]{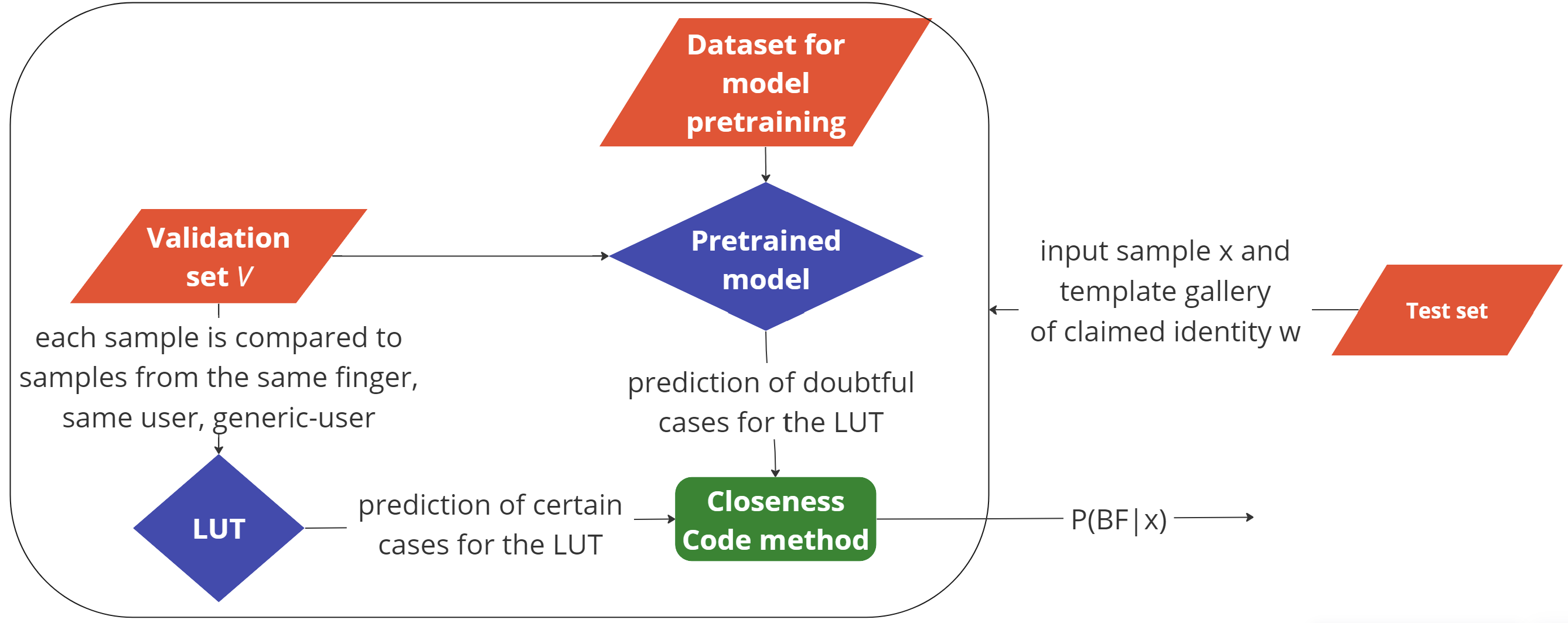}
    \caption{Experimental protocol to evaluate the effectiveness of using the CC module in increasing the performance of a PAD. The test set samples are classified both by the pre-trained model and by the CC method, which uses not only the validation set V but also the bona-fide samples of the same user as a template gallery for calculating the closeness binary code.}
    \label{fig:schemaprotoc}
\end{figure}

Both experimental protocols are schematized in Fig. \ref{fig:schemaprotoc}: each test sample is classified by the pre-trained model and the LUT. To obtain the latter classification, in addition to the input sample, the \textit{bona fide} samples of the claimed ID are also passed to the method as a template gallery, which is combined with the validation data for calculating the CC through hierarchical distances. 

This protocol allows us to evaluate the CC as an add-on module that can be added later to a pre-trained PAD. The experimental evaluation is set to highlight whether the addition of the CC module benefits the functioning of the PAD, especially when integrated into an AFIS.

To better analyze the utility of the CC in different scenarios (for example, when the PAD is attacked with never-seen-before PAIs), we evaluated different combinations of LivDet data sets as validation and test sets. In particular, the datasets involved in the analysis are LivDet 2017, LivDet 2019 and LivDet 2021 \cite{micheletto2023review}, whose compositions are shown in Table \ref{table:datasetcomposition}. It is worth noting that this protocol allows us to evaluate the ability of the method to generalize on PAI materials and generation techniques unknown in the training phase.

Following the ISO/IEC SC37 standards \cite{iso}, we evaluated the performance of the PAD systems using two primary metrics: the Bona Fide Presentation Classification Error Rate (BPCER) and the Attack Presentation Classification Error Rate (APCER). BPCER measures the proportion of bona fide presentations incorrectly classified as attacks, while APCER quantifies the proportion of attack presentations misclassified as bona fide. Additionally, we computed the Average Classification Error Rate (ACER) to provide a balanced assessment and facilitate comparison with state-of-the-art methods in the literature. The ACER is defined as the arithmetic mean of APCER and BPCER.


\begin{table*}[]
\caption{Composition of the datasets used in the experimental evaluation of the CC method \cite{micheletto2023review}.}
\label{table:datasetcomposition}
\centering
\resizebox{0.55\textwidth}{!}{
\begin{tabular}{|c|c|c|c|c|c|c|}
\hline
\multirow{3}{*}{\textbf{Dataset}}   & \multirow{3}{*}{\textbf{\begin{tabular}[c]{@{}c@{}}LivDet 2019 \\ Training \\ {[}train19{]}\end{tabular}}} & \multirow{3}{*}{\textbf{\begin{tabular}[c]{@{}c@{}}LivDet 2021 \\ Training \\ {[}train21{]}\end{tabular}}} & \multirow{3}{*}{\textbf{\begin{tabular}[c]{@{}c@{}}LivDet 2017 \\ Test \\ {[}test17{]}\end{tabular}}} & \multirow{3}{*}{\textbf{\begin{tabular}[c]{@{}c@{}}LivDet 2019 \\ Test\\ {[}test19{]}\end{tabular}}} & \multirow{3}{*}{\textbf{\begin{tabular}[c]{@{}c@{}}LivDet 2021 \\ Test Consensual\\ {[}test21c{]}\end{tabular}}} & \multirow{3}{*}{\textbf{\begin{tabular}[c]{@{}c@{}}LivDet 2021 \\ Test ScreenSpoof\\ {[}test21s{]}\end{tabular}}} \\
&&&&&&\\
&&&&&&\\ \hline
\textbf{BF {[}L{]}}               & 1000& 1250& 1700& 1020& 2050& 2050\\ \hline
\textbf{Latex{[}LX{]}}              & -& 750& 680& -& -& -\\ \hline
\textbf{RProFast{[}RPF{]}}          & -& 750& -& -& -& -\\ \hline
\textbf{Woodglue {[}WG{]}}          & 400& -& -& -& -& -\\ \hline
\textbf{BodyDouble {[}BD{]}}        & 400& -& -& -& 820& 820\\ \hline
\textbf{Ecoflex {[}ECO{]}}          & 400& -& -& -& -& -\\ \hline
\textbf{Liquid Ecoflex {[}L.ECO{]}} & -& -& 680& 408& -& -\\ \hline
\textbf{Gelatine {[}GEL{]}}         & -& -& 680& -& -& -\\ \hline
\textbf{Mix1 {[}M1{]}}              & -& -& -& 408& 820& 820\\ \hline
\textbf{Mix2 {[}M2{]}}              & -& -& -& 408& -& -\\ \hline
\textbf{Elmers Glue {[}EG{]}}       & -& -& -& -& 820& 820\\ \hline
\end{tabular}}
\end{table*}

\subsection{Experimental results}
This section presents the experimental evaluation results aimed at investigating the effective support of the CC module on any white-box or black-box PAD regardless of its implementation characteristics and nature, whether hand-crafted or deep-learning-based.
In particular, the evaluation aims to highlight the impact of adding the CC module in relation to worst-case applications, such as detecting unknown PAIs. The section, therefore, first presents a preliminary analysis that allows us to confirm the hypothesis underlying the method, i.e., the user-specific hierarchical proximity, and then moves on to the evaluation of the actual add-on on different types of PADs, both standalone and integrated into an AFIS. To provide a comprehensive benchmark and position our approach within the broader context of state-of-the-art PAD systems, Table \ref{table:sota_err} presents the baseline performances of the evaluated PAD methods alongside other established methods from the literature. Establishing a clear reference point allows us to highlight the improvements enabled by the CC module, demonstrating its capacity to elevate PAD performance to competitive levels, as shown later in Section \ref{sec:impact}.

\begin{table*}[]
\caption{ACER, APCER and BPCER of the PAD methods selected for analysis and other state-of-the-art PADs. The values in parentheses in the \textit{SpoofBuster} column represent the Failure to Extract (FTX) rate.}
\label{table:sota_err}
\centering
\resizebox{0.85\textwidth}{!}{
\begin{tabular}{|cc|c|c|c|c|c|c|c|c|c|c|} \hline 

\multirow{2}{*}{}  & & \multicolumn{5}{c|}{\textbf{Investigated baseline methods}}& \multicolumn{5}{c|}{\textbf{State-of-the-art methods}}\\ \cline{3-12}

& & \textbf{\begin{tabular}[c]{@{}c@{}}PADUnk21\\ \cite{gonzalez2021local}\end{tabular}} & \textbf{MEGVII} & \textbf{CNN} & \textbf{\begin{tabular}[c]{@{}c@{}}BSIF\\ \cite{bsifghiani}\end{tabular}} & \textbf{\begin{tabular}[c]{@{}c@{}}LBP\\ \cite{lbpfing}\end{tabular}} & \textbf{\begin{tabular}[c]{@{}c@{}}SpoofBuster\\ \cite{chugh2018fingerprint}\end{tabular}} & \textbf{\begin{tabular}[c]{@{}c@{}}Slim-ResCNN\\ \cite{slim}\end{tabular}} & \textbf{\begin{tabular}[c]{@{}c@{}}PADUnk19\\ \cite{gonzalez2021fingerprint}\end{tabular}} & \textbf{\begin{tabular}[c]{@{}c@{}}sDSIFT\\ \cite{contreras}\end{tabular}} & \textbf{\begin{tabular}[c]{@{}c@{}}PADRD-Net\\ \cite{fei2024fingerprint}\end{tabular}} \\ \hline \multicolumn{1}{|c|}{\multirow{3}{*}{\textbf{test2019}}} & \textbf{APCER} & 18.46 & 0.49 & 7.92 & 8.17 & 15.03 & 0.08 & 1.14 & 1.55 & 12.91 & - \\ \cline{2-12} \multicolumn{1}{|c|}{} & \textbf{BPCER} & 0.78 & 1.27 & 4.41 & 1.27 & 1.37 & 0.49 & 0.39 & 3.24 & 1.96 & - \\ \cline{2-12} \multicolumn{1}{|c|}{} & \textbf{ACER} & 9.62 & 0.88 & 6.17 & 4.72 & 8.20 & 0.29 & 0.77 & 2.40 & 7.44 & 1.40 \\ \hline \multicolumn{1}{|c|}{\multirow{3}{*}{\textbf{test2021}}} & \textbf{APCER} & 36.2 & 2.72 & 13.41 & 19.27 & 24.19 & 0.41 (6.83) & 0.41 & 9.27 & 3.94 & - \\ \cline{2-12} \multicolumn{1}{|c|}{} & \textbf{BPCER} & 1.46 & 0.05 & 9.66 & 4.49 & 8.49 & 1.02 (8.39) & 0.39 & 6.24 & 8.98 & - \\ \cline{2-12} \multicolumn{1}{|c|}{} & \textbf{ACER} & 18.83 & 1.39 & 11.54 & 11.88 & 16.34 & 0.72 (7.61) & 0.40 & 7.76 & 6.46 & 4.70 \\ \hline \multicolumn{1}{|c|}{\multirow{3}{*}{\textbf{test2021ss}}} & \textbf{APCER} & 18.41 & 13.62 & 82.76 & 53.29 & 24.19 & 0.41 (14.92) & 5.77 & 16.06 & 26.67 & - \\ \cline{2-12} \multicolumn{1}{|c|}{} & \textbf{BPCER} & 1.46 & 0.05 & 9.66 & 4.49 & 8.49 & 1.02 (8.39) & 0.39 & 6.24 & 8.98 & - \\ \cline{2-12} \multicolumn{1}{|c|}{} & \textbf{ACER} & 9.94 & 6.84 & 46.21 & 28.89 & 16.34 & 0.72 (11.66) & 3.08 & 11.15 & 17.82 & 7.14 \\ \hline 
\end{tabular}}
\end{table*}

\subsection{Hierarchical distance analysis}
\label{sec:hieran}
To pave the way for a detailed examination of our results in this experimental section, we first present an analysis conducted on hierarchical distances for each detector.
In fact, the CC method can be effectively applied only if the hypothesis of proximity between BF samples of the same user is verified.
For this reason, for each of the experimental evaluation datasets and each PAD feature representation, we evaluated the percentage of BF and PA samples geometrically close to a BF sample from the same finger, user, or generic.
We then averaged and reported the results in Table \ref{table:hdist}.

\begin{figure}[]

   \centering
      \subfigure[]{\includegraphics[width=0.48\textwidth]{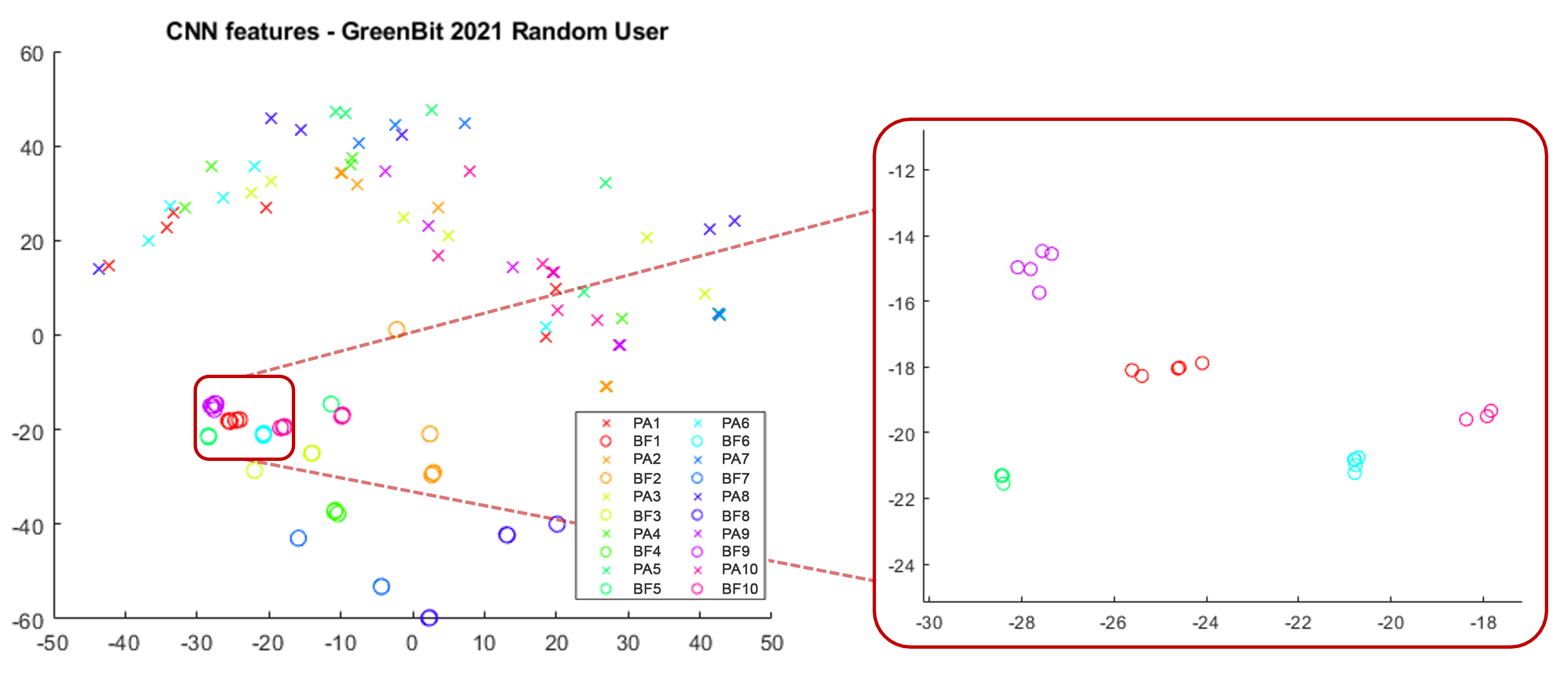}}
     
\caption{t-SNE visualization of feature embeddings for a random user from GreenBit 2021 dataset using the CNN PADs. Each color represent a different finger of the user.}
\label{fig:zoom}
\vspace{-10pt}
\end{figure}

\begin{figure}[]

   \centering
      \subfigure[]{\includegraphics[width=0.35\textwidth]{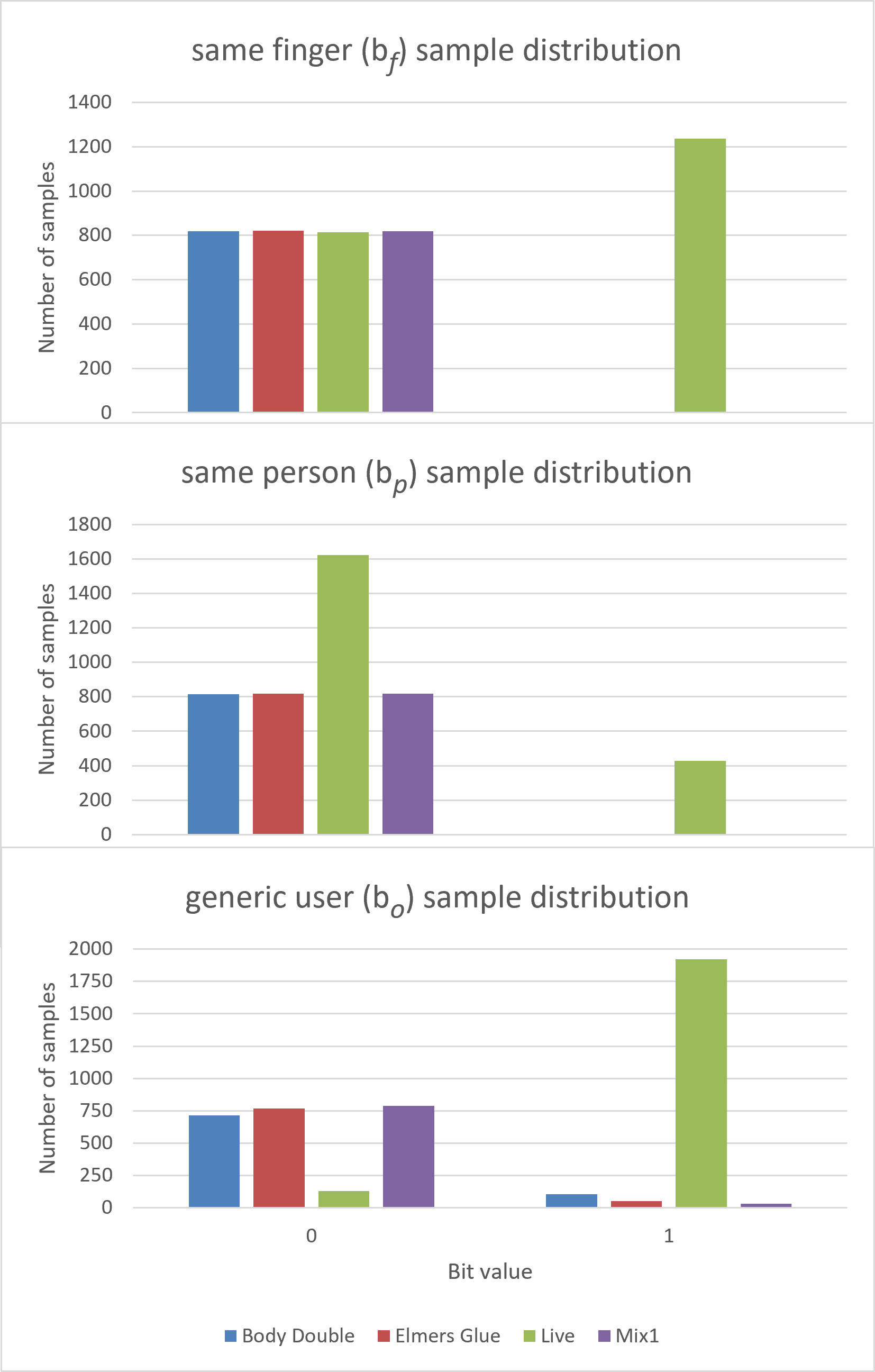}}
     
\caption{Closeness code feature distribution from the LivDet 2021 validation set for the CNN PAD.}
\label{fig:histogram}
\vspace{-10pt}
\end{figure}

These data offer an insightful perspective into the hierarchical distances of each PAD, giving substance to the concept of hierarchical proximity. By hierarchical proximity, we refer to the propensity of biometric features from the same finger or person ($b_f=1$ or $b_p=1$, respectively) to cluster together in the feature space. Additionally, we consider proximity to any generic \textit{bona fide} fingerprint ($b_o=1$). It is clear from the table that BF fingerprints uphold a certain degree of this hierarchical proximity. The degree varies, however, depending on the specific feature representation and the hierarchical level.
For instance, the PADUnk21 detector shows the highest percentage of BF samples being close to other acquisitions of the same finger (95.81\%), which diminishes if we consider samples from the same user (44.98\%) and reaches the lowest value (52.65\%) when considering the proximity to any generic \textit{bona fide} fingerprint. These patterns fluctuate for the different PADs but remain considerably high, reinforcing the solid hierarchical closeness among \textit{bona fide} fingerprints. To further illustrate the concept of hierarchical proximity, Figure \ref{fig:zoom} shows the 2D t-SNE \cite{van2008visualizing} plot of CNN feature embeddings for a randomly selected user from the LivDet 2021 validation set. The feature space reveals close clustering of different acquisitions of the same fingers, each color-coded per finger. This indicates that BF samples from the same finger maintain significant proximity to each other while also showing proximity among samples from other fingers of the same user, thereby illustrating the hierarchical nature of the proximity. Contrarily, the PA samples do not exhibit such clustering behavior and are randomly distributed in the feature space. This is consistent across all PADs, as shown in Table \ref{table:hdist}, where the percentages of PA samples close to a BF are significantly lower, emphasizing the absence of user-specific hierarchical proximity for PA fingerprints.
However, it is notable that depending on the PAD, a fair percentage of PA fingerprints can be found close to a generic BF fingerprint ($b_o=1$). For instance, this is more apparent for the CNN detector (34.04\%). The histogram in Figure \ref{fig:histogram} supports these findings by showing the distribution of samples for the CNN detector on LivDet 2021. It is evident that there are no PA samples close to the BF samples of the same finger and person, whereas some PA samples appear close to generic \textit{bona fide} samples, particularly for those fabricated with the "Elmers Glue" material. This is typical of less accurate PADs or in the presence of unknown attacks. In other words, when the PA and BF distributions are not perfectly separated, the percentage of PAs close to BF samples increases.
Our goal is to confront these cases directly by leveraging the potential of our proposed CC method. Acknowledging that overlaps between the distributions can occur, our primary focus remains to ensure that the principle of user-specific hierarchical distance does not apply to PA fingerprints.
As we will demonstrate in the following section, the Closeness Binary Code provides valuable insights in these demanding scenarios, assisting the PAD system to better distinguish between \textit{bona fide} and attack presentations.

\begin{table}[]
\caption{Mean hierarchical distances of each sample with the rest of the set, obtained by averaging the distances from each validation set of the experimental evaluation.}
\centering
\label{table:hdist}
\resizebox{0.45\textwidth}{!}{
\begin{tabular}{c|ccc|ccc|}
\cline{2-7}
\multicolumn{1}{l|}{}                 & \multicolumn{3}{c|}{\textbf{\% BF close to a BF}}                                       & \multicolumn{3}{c|}{\textbf{\% PA close to a BF}}                                       \\ \cline{2-7} 
\multicolumn{1}{l|}{}                 & \multicolumn{1}{c|}{\textbf{$b_f$=1}} & \multicolumn{1}{c|}{\textbf{$b_p$=1}} & \textbf{$b_o$=1} & \multicolumn{1}{c|}{\textbf{$b_f$=1}} & \multicolumn{1}{c|}{\textbf{$b_p$=1}} & \textbf{$b_o$=1} \\ \hline
\multicolumn{1}{|c|}{\textbf{PADUnk21}} & \multicolumn{1}{c|}{95.81}         & \multicolumn{1}{c|}{44.98}         & 52.65         & \multicolumn{1}{c|}{2.22}          & \multicolumn{1}{c|}{2.42}          & 14.43         \\ \hline
\multicolumn{1}{|c|}{\textbf{MEGVII}} & \multicolumn{1}{c|}{44.38}         & \multicolumn{1}{c|}{48.19}         & 98.80         & \multicolumn{1}{c|}{0.03}          & \multicolumn{1}{c|}{0.23}          & 5.70          \\ \hline
\multicolumn{1}{|c|}{\textbf{CNN}} & \multicolumn{1}{c|}{62.31}         & \multicolumn{1}{c|}{24.62}         & 88.70         & \multicolumn{1}{c|}{0.41}          & \multicolumn{1}{c|}{1.64}          & 34.04         \\ \hline
\multicolumn{1}{|c|}{\textbf{BSIF}}   & \multicolumn{1}{c|}{83.26}         & \multicolumn{1}{c|}{63.95}         & 82.61         & \multicolumn{1}{c|}{0.85}          & \multicolumn{1}{c|}{2.23}          & 24.03         \\ \hline
\end{tabular}}
\end{table}

\subsection{Impact of the CC module on PAD results}
\label{sec:impact}

Evaluating the Closeness Binary Code module's effectiveness is a crucial component of our investigation. Tables \ref{table:TabBSIF}-\ref{table:TabPADUnk21}, outlining the performance of PAD systems, both with and without the CC module, facilitate a comprehensive understanding of its impact on the biometric security framework. Specifically, these results underscore the dual impact of the CC module in reducing the BPCER and the APCER. The variable levels of enhancement indicate that its success partially relies on the specificities of the PAD method and the dataset at hand.

\begin{figure}[]

   \centering
      \subfigure[]{\includegraphics[width=0.44\textwidth]{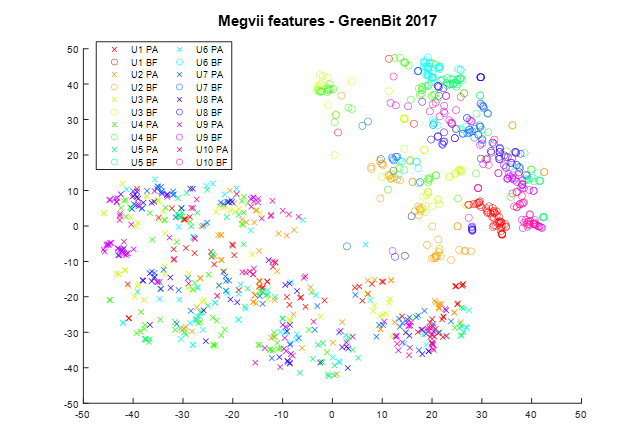}}
     \subfigure[]{\includegraphics[width=0.44\textwidth]{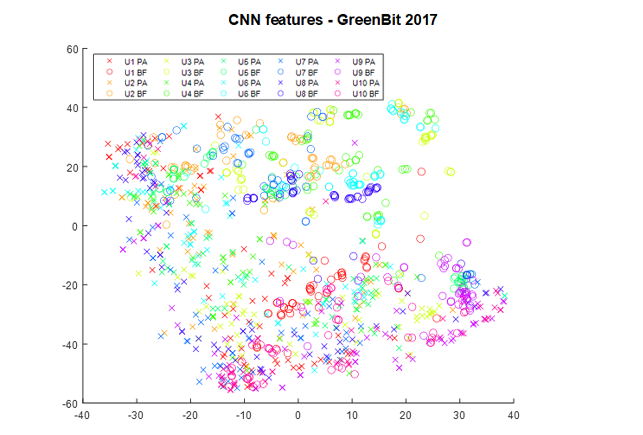}}
     
\caption{t-SNE visualization of feature embeddings for 10 random users from GreenBit 2017 dataset using the MEGVII (a) and CNN (b) PADs.}
\label{fig:tsne1}
\vspace{-10pt}
\end{figure}

\begin{table}
\centering
\caption{Performance comparison of BSIF without and with CC module for different validation and test sets (the material in common between the two sets is shown in brackets). The last two columns report the performance decrease $\Delta$ obtained with the CC module. Lower overall errors (summing increase of APCER and BPCER) are shown in green, higher overall errors in red.}
\label{table:TabBSIF}
\resizebox{\linewidth}{!}{
\begin{tabular}{|c|c|c|c|c|c|c|c|} 
\hline
\multirow{2}{*}{\textbf{\begin{tabular}[c]{@{}c@{}}Val.\\ set\end{tabular}}} & \multirow{2}{*}{\textbf{Test~set}} & \multicolumn{2}{c|}{\textbf{BSIF }} & \multicolumn{2}{c|}{\textbf{BSIF + CC }} & \multirow{2}{*}{\textbf{\begin{tabular}[c]{@{}c@{}}$\Delta$\\ BPCER\end{tabular} }} & \multirow{2}{*}{\textbf{\begin{tabular}[c]{@{}c@{}}$\Delta$\\ APCER\end{tabular} }} \\ 
\cline{3-6}
 &  & \textbf{BPCER} & \textbf{APCER} & \textbf{BPCER} & \textbf{APCER} &  &  \\ 
\hline
\multirow{3}{*}{\textbf{test21 }} & \textbf{test17 [-]} & 2.88 & 20.54 & 1.18 & 8.68 & \textcolor[rgb]{0,0.502,0}{-1.7} & \textcolor[rgb]{0,0.502,0}{-11.86} \\ 
\cline{2-8}
 & \textbf{test19 [M1]} & 1.27 & 8.17 & 1.27 & 2.45 & \textcolor[rgb]{0,0.502,0}{0} & \textcolor[rgb]{0,0.502,0}{-5.72} \\ 
\cline{2-8}
 & \textbf{train19 [BD]} & 1.7 & 9.42 & 1.3 & 4.17 & \textcolor[rgb]{0,0.502,0}{-0.4} & \textcolor[rgb]{0,0.502,0}{-5.25} \\ 
\hline
\multirow{4}{*}{\textbf{test19 }} & \textbf{test17 [LE]} & 2.88 & 20.54 & 0.82 & 17.4 & \textcolor[rgb]{0,0.502,0}{-2.06} & \textcolor[rgb]{0,0.502,0}{-3.14} \\ 
\cline{2-8}
 & \textbf{test21 [M1]} & 4.49 & 19.27 & 0.59 & 6.14 & \textcolor[rgb]{0,0.502,0}{-3.9} & \textcolor[rgb]{0,0.502,0}{-13.13} \\ 
\cline{2-8}
 & \textbf{test21ss [M1]} & 4.49 & 53.29 & 0.59 & 24.02 & \textcolor[rgb]{0,0.502,0}{-3.9} & \textcolor[rgb]{0,0.502,0}{-29.27} \\ 
\cline{2-8}
 & \textbf{train19 [-]} & 1.7 & 9.42 & 0.7 & 6.33 & \textcolor[rgb]{0,0.502,0}{-1} & \textcolor[rgb]{0,0.502,0}{-3.09} \\ 
\hline
\multirow{4}{*}{\textbf{test17 }} & \textbf{test19 [LE]} & 1.27 & 8.17 & 1.47 & 1.39 & \textcolor[rgb]{0,0.502,0}{0.2} & \textcolor[rgb]{0,0.502,0}{-6.78} \\ 
\cline{2-8}
 & \textbf{test21 [-]} & 4.49 & 19.27 & 1.66 & 10.33 & \textcolor[rgb]{0,0.502,0}{-2.83} & \textcolor[rgb]{0,0.502,0}{-8.94} \\ 
\cline{2-8}
 & \textbf{test21ss [-]} & 4.49 & 53.29 & 1.66 & 23.82 & \textcolor[rgb]{0,0.502,0}{-2.83} & \textcolor[rgb]{0,0.502,0}{-29.47} \\ 
\cline{2-8}
 & \textbf{train19 [-]} & 1.7 & 9.42 & 0.6 & 1.67 & \textcolor[rgb]{0,0.502,0}{-1.1} & \textcolor[rgb]{0,0.502,0}{-7.75} \\ 
\hline
\multirow{4}{*}{\textbf{train19 }} & \textbf{test17 [-]} & 2.88 & 20.54 & 1.59 & 2.35 & \textcolor[rgb]{0,0.502,0}{-1.29} & \textcolor[rgb]{0,0.502,0}{-18.19} \\ 
\cline{2-8}
 & \textbf{test19 [-]} & 1.27 & 8.17 & 1.76 & 1.39 & \textcolor[rgb]{0,0.502,0}{0.49} & \textcolor[rgb]{0,0.502,0}{-6.78} \\ 
\cline{2-8}
 & \textbf{test21 [BD]} & 4.49 & 19.27 & 2.24 & 4.59 & \textcolor[rgb]{0,0.502,0}{-2.25} & \textcolor[rgb]{0,0.502,0}{-14.68} \\ 
\cline{2-8}
 & \textbf{test21ss [BD]} & 4.49 & 53.29 & 2.24 & 13.74 & \textcolor[rgb]{0,0.502,0}{-2.25} & \textcolor[rgb]{0,0.502,0}{-39.55} \\
\hline
\end{tabular}}
\end{table}

\begin{table}
\centering
\caption{Performance comparison of CNN without and with CC module for different validation and test sets (the material in common between the two sets is shown in brackets). The last two columns report the performance decrease $\Delta$ obtained with the CC module. Lower overall errors (summing increase of APCER and BPCER) are shown in green, higher overall errors in red.}
\label{table:TabCNN}
\resizebox{\linewidth}{!}{
\begin{tabular}{|c|c|c|c|c|c|c|c|} 
\hline
\multirow{2}{*}{\textbf{\begin{tabular}[c]{@{}c@{}}Val.\\ set\end{tabular}}} & \multirow{2}{*}{\textbf{Test~set}} & \multicolumn{2}{c|}{\textbf{CNN}} & \multicolumn{2}{c|}{\textbf{CNN+ CC}} & \multirow{2}{*}{\textbf{\begin{tabular}[c]{@{}c@{}}$\Delta$\\ BPCER\end{tabular} }} & \multirow{2}{*}{\textbf{\begin{tabular}[c]{@{}c@{}}$\Delta$\\ APCER\end{tabular} }} \\ 
\cline{3-6}
 &  & \textbf{BPCER} & \textbf{APCER} & \textbf{BPCER} & \textbf{APCER} &  &  \\ 
\hline
\multirow{3}{*}{\textbf{test21 }} & \textbf{test17 [-]} &14,12&24,46&6,41&20,39& \textcolor[rgb]{0,0.502,0}{-7,71}& \textcolor[rgb]{0,0.502,0}{-4,07} \\			
\cline{2-8}			
 & \textbf{test19 [M1]} &4,41&7,92&3,14&4,9& \textcolor[rgb]{0,0.502,0}{-1,27}& \textcolor[rgb]{0,0.502,0}{-3,02} \\			
\cline{2-8}			
 & \textbf{train19 [BD]} &10,7&24,33&9,3&22,83& \textcolor[rgb]{0,0.502,0}{-1,4}& \textcolor[rgb]{0,0.502,0}{-1,5} \\			
\hline			
\multirow{4}{*}{\textbf{test19 }} & \textbf{test17 [LE]} &14,12&24,46&5,65&22,55& \textcolor[rgb]{0,0.502,0}{-8,47}& \textcolor[rgb]{0,0.502,0}{-1,91} \\			
\cline{2-8}			
 & \textbf{test21 [M1]} &9,66&13,41&5,22&3,62& \textcolor[rgb]{0,0.502,0}{-4,44}& \textcolor[rgb]{0,0.502,0}{-9,79} \\			
\cline{2-8}			
 & \textbf{test21ss [M1]} &9,66&82,76&5,22&78,09& \textcolor[rgb]{0,0.502,0}{-4,44}& \textcolor[rgb]{0,0.502,0}{-4,67} \\			
\cline{2-8}			
 & \textbf{train19 [-]} &10,7&24,33&6,90&22,5& \textcolor[rgb]{0,0.502,0}{-3,8}& \textcolor[rgb]{0,0.502,0}{-1,83} \\			
\hline			
\multirow{4}{*}{\textbf{test17 }} & \textbf{test19 [LE]} &4,41&7,92&3,73&3,02& \textcolor[rgb]{0,0.502,0}{-0,68}& \textcolor[rgb]{0,0.502,0}{-4,9} \\			
\cline{2-8}			
 & \textbf{test21 [-]} &9,66&13,41&7,51&7,56& \textcolor[rgb]{0,0.502,0}{-2,15}& \textcolor[rgb]{0,0.502,0}{-5,85} \\			
\cline{2-8}			
 & \textbf{test21ss [-]} &9,66&82,76&7,51&61,54& \textcolor[rgb]{0,0.502,0}{-2,15}& \textcolor[rgb]{0,0.502,0}{-21,22} \\			
\cline{2-8}			
 & \textbf{train19 [-]} &10,7&24,33&10,1&12,08& \textcolor[rgb]{0,0.502,0}{-0,6}& \textcolor[rgb]{0,0.502,0}{-12,25} \\			
\hline			
\multirow{4}{*}{\textbf{train19 }} & \textbf{test17 [-]} &14,12&24,46&8,59&13,28& \textcolor[rgb]{0,0.502,0}{-5,53}& \textcolor[rgb]{0,0.502,0}{-11,18} \\			
\cline{2-8}			
 & \textbf{test19 [-]} &4,41&7,92&2,06&3,02& \textcolor[rgb]{0,0.502,0}{-2,35}& \textcolor[rgb]{0,0.502,0}{-4,9} \\			
\cline{2-8}			
 & \textbf{test21 [BD]} &9,66&13,41&5,95&12,24& \textcolor[rgb]{0,0.502,0}{-3,71}& \textcolor[rgb]{0,0.502,0}{-1,17} \\			
\cline{2-8}			
 & \textbf{test21ss [BD]} &9,66&82,76&5,95&59,8& \textcolor[rgb]{0,0.502,0}{-3,71}& \textcolor[rgb]{0,0.502,0}{-22,96} \\			

\hline
\end{tabular}}
\end{table}

\begin{table}
\centering
\caption{Performance comparison of MEGVII without and with CC module for different validation and test sets (the material in common between the two sets is shown in brackets). The last two columns report the performance decrease $\Delta$ obtained with the CC module. Lower overall errors (summing increase of APCER and BPCER) are shown in green, and higher overall errors are in red.}
\label{table:TabMEG}
\resizebox{\linewidth}{!}{
\begin{tabular}{|c|c|c|c|c|c|c|c|} 
\hline
\multirow{2}{*}{\textbf{\begin{tabular}[c]{@{}c@{}}Val.\\ set\end{tabular}}} & \multirow{2}{*}{\textbf{Test~set}} & \multicolumn{2}{c|}{\textbf{MEGVII}} & \multicolumn{2}{c|}{\textbf{MEGVII+ CC}} & \multirow{2}{*}{\textbf{\begin{tabular}[c]{@{}c@{}}$\Delta$\\ BPCER\end{tabular} }} & \multirow{2}{*}{\textbf{\begin{tabular}[c]{@{}c@{}}$\Delta$\\ APCER\end{tabular} }} \\ 
\cline{3-6}
 &  & \textbf{BPCER} & \textbf{APCER} & \textbf{BPCER} & \textbf{APCER} &  &  \\ 
\hline
\multirow{3}{*}{\textbf{test21 }} & \textbf{test17 [-]} & 2.24 & 0.88 & 0.35 & 2.6 & \textcolor[rgb]{0,0.502,0}{-1.89} & \textcolor[rgb]{0,0.502,0}{1.72} \\ 
\cline{2-8}
 & \textbf{test19 [M1]} & 1.27 & 0.49 & 0.49 & 0.49 & \textcolor[rgb]{0,0.502,0}{-0.78} & \textcolor[rgb]{0,0.502,0}{0.00} \\ 
\cline{2-8}
 & \textbf{train19 [BD]} & 0.6 & 0 & 0.7 & 0.08 & \textcolor{red}{0.1} & \textcolor{red}{0.08} \\ 
\hline
\multirow{4}{*}{\textbf{test19 }} & \textbf{test17 [LE]} & 2.24 & 0.88 & 0.76 & 1.18 & \textcolor[rgb]{0,0.502,0}{-1.48} & \textcolor[rgb]{0,0.502,0}{0.3} \\ 
\cline{2-8}
 & \textbf{test21 [M1]} & 0.05 & 2.72 & 0.05 & 0.2 & \textcolor[rgb]{0,0.502,0}{0} & \textcolor[rgb]{0,0.502,0}{-2.52} \\ 
\cline{2-8}
 & \textbf{test21ss [M1]} & 0.05 & 13.62 & 0.05 & 8.78 & \textcolor[rgb]{0,0.502,0}{0} & \textcolor[rgb]{0,0.502,0}{-4.84} \\ 
\cline{2-8}
 & \textbf{train19 [-]} & 0.6 & 0 & 0.5 & 0.17 & \textcolor{red}{-0.1} & \textcolor{red}{0.17} \\ 
\hline
\multirow{4}{*}{\textbf{test17 }} & \textbf{test19 [LE]} & 1.27 & 0.49 & 0.2 & 0.16 & \textcolor[rgb]{0,0.502,0}{-1.07} & \textcolor[rgb]{0,0.502,0}{-0.33} \\ 
\cline{2-8}
 & \textbf{test21 [-]} & 0.05 & 2.72 & 0.24 & 1.75 & \textcolor[rgb]{0,0.502,0}{0.19} & \textcolor[rgb]{0,0.502,0}{-0.97} \\ 
\cline{2-8}
 & \textbf{test21ss [-]} & 0.05 & 13.62 & 0.24 & 8.98 & \textcolor[rgb]{0,0.502,0}{0.19} & \textcolor[rgb]{0,0.502,0}{-4.64} \\ 
\cline{2-8}
 & \textbf{train19 [-]} & 0.6 & 0 & 0.8 & 0 & \textcolor{red}{0.2} & \textcolor{red}{0.00} \\ 
\hline
\multirow{4}{*}{\textbf{train19 }} & \textbf{test17 [-]} & 2.24 & 0.88 & 0.88 & 0.83 & \textcolor[rgb]{0,0.502,0}{-1.36} & \textcolor[rgb]{0,0.502,0}{-0.05} \\ 
\cline{2-8}
 & \textbf{test19 [-]} & 1.27 & 0.49 & 0.1 & 0.16 & \textcolor[rgb]{0,0.502,0}{-1.17} & \textcolor[rgb]{0,0.502,0}{-0.33} \\ 
\cline{2-8}
 & \textbf{test21 [BD]} & 0.05 & 2.72 & 0.05 & 2.36 & \textcolor[rgb]{0,0.502,0}{0.00} & \textcolor[rgb]{0,0.502,0}{-0.36} \\ 
\cline{2-8}
 & \textbf{test21ss [BD]} & 0.05 & 13.62 & 0.05 & 12.11 & \textcolor[rgb]{0,0.502,0}{0.00} & \textcolor[rgb]{0,0.502,0}{-1.51} \\
\hline
\end{tabular}}
\end{table}

\begin{table}
\centering
\caption{Performance comparison of PADUnk21 without and with CC module for different validation and test sets (the material in common between the two sets is shown in brackets). The last two columns report the performance decrease $\Delta$ obtained with the CC module. Lower overall errors (summing increase of APCER and BPCER) are shown in green, and higher overall errors are in red.}
\label{table:TabPADUnk21}
\resizebox{\linewidth}{!}{
\begin{tabular}{|c|c|c|c|c|c|c|c|} 
\hline
\multirow{2}{*}{\textbf{\begin{tabular}[c]{@{}c@{}}Val.\\ set\end{tabular}}} & \multirow{2}{*}{\textbf{Test~set}} & \multicolumn{2}{c|}{\textbf{PADUnk21}} & \multicolumn{2}{c|}{\textbf{PADUnk21+ CC}} & \multirow{2}{*}{\textbf{\begin{tabular}[c]{@{}c@{}}$\Delta$\\ BPCER\end{tabular} }} & \multirow{2}{*}{\textbf{\begin{tabular}[c]{@{}c@{}}$\Delta$\\ APCER\end{tabular} }} \\ 
\cline{3-6}
 &  & \textbf{BPCER} & \textbf{APCER} & \textbf{BPCER} & \textbf{APCER} &  &  \\ 
\hline
\multirow{3}{*}{\textbf{test21 }} & \textbf{test17 [-]} & 4.76 & 28.92 & 4.94 & 9.9 & \textcolor[rgb]{0,0.502,0}{0.18} & \textcolor[rgb]{0,0.502,0}{-19.02} \\ 
\cline{2-8}
 & \textbf{test19 [M1]} & 0.78 & 18.46 & 3.53 & 0.82 & \textcolor[rgb]{0,0.502,0}{2.75} & \textcolor[rgb]{0,0.502,0}{-17.64} \\ 
\cline{2-8}
 & \textbf{train19 [BD]} & 1.7 & 30.17 & 5.6 & 6.58 & \textcolor[rgb]{0,0.502,0}{3.9} & \textcolor[rgb]{0,0.502,0}{-23.59} \\ 
\hline
\multirow{4}{*}{\textbf{test19 }} & \textbf{test17 [LE]} & 4.76 & 28.92 & 6.53 & 5.49 & \textcolor[rgb]{0,0.502,0}{1.77} & \textcolor[rgb]{0,0.502,0}{-23.43} \\ 
\cline{2-8}
 & \textbf{test21 [M1]} & 1.46 & 37.2 & 3.27 & 0.73 & \textcolor[rgb]{0,0.502,0}{1.81} & \textcolor[rgb]{0,0.502,0}{-36.47} \\ 
\cline{2-8}
 & \textbf{test21ss [M1]} & 1.46 & 18.41 & 3.27 & 5.16 & \textcolor[rgb]{0,0.502,0}{1.81} & \textcolor[rgb]{0,0.502,0}{-13.25} \\ 
\cline{2-8}
 & \textbf{train19 [-]} & 1.7 & 30.17 & 3.3 & 4.17 & \textcolor[rgb]{0,0.502,0}{1.6} & \textcolor[rgb]{0,0.502,0}{-26} \\ 
\hline
\multirow{4}{*}{\textbf{test17 }} & \textbf{test19 [LE]} & 0.78 & 18.46 & 1.96 & 3.43 & \textcolor[rgb]{0,0.502,0}{1.18} & \textcolor[rgb]{0,0.502,0}{-15.03} \\ 
\cline{2-8}
 & \textbf{test21 [-]} & 1.46 & 37.2 & 1.07 & 8.54 & \textcolor[rgb]{0,0.502,0}{-0.39} & \textcolor[rgb]{0,0.502,0}{-28.66} \\ 
\cline{2-8}
 & \textbf{test21ss [-]} & 1.46 & 18.41 & 1.07 & 3.05 & \textcolor[rgb]{0,0.502,0}{-0.39} & \textcolor[rgb]{0,0.502,0}{-15.36} \\ 
\cline{2-8}
 & \textbf{train19 [-]} & 1.7 & 30.17 & 0.9 & 2.17 & \textcolor[rgb]{0,0.502,0}{-0.8} & \textcolor[rgb]{0,0.502,0}{-28.00} \\ 
\hline
\multirow{4}{*}{\textbf{train19 }} & \textbf{test17 [-]} & 4.76 & 28.92 & 7.41 & 0.59 & \textcolor[rgb]{0,0.502,0}{2.65} & \textcolor[rgb]{0,0.502,0}{-28.33} \\ 
\cline{2-8}
 & \textbf{test19 [-]} & 0.78 & 18.46 & 2.45 & 0.74 & \textcolor[rgb]{0,0.502,0}{1.67} & \textcolor[rgb]{0,0.502,0}{-17.72} \\ 
\cline{2-8}
 & \textbf{test21 [BD]} & 1.46 & 37.2 & 2.39 & 0.65 & \textcolor[rgb]{0,0.502,0}{0.93} & \textcolor[rgb]{0,0.502,0}{-36.55} \\ 
\cline{2-8}
 & \textbf{test21ss [BD]} & 1.46 & 18.41 & 2.39 & 1.22 & \textcolor[rgb]{0,0.502,0}{0.93} & \textcolor[rgb]{0,0.502,0}{-17.19} \\
\hline
\end{tabular}}
\end{table}

For instance, the influence of the CC module is strikingly evident in the PADUnk21 system. Although it is the only detector that showed a slight increase in BPCER on average, the counterbalancing effect is a substantial decrease in APCER, varying from a minimum of 13.25\% to a notable maximum of 36.55\%. In this case, the CC module operates as a pivotal ``balancer'', counteracting PADUnk21's inherent weaknesses in accurately distinguishing presentation attacks.
A similar impact can also be observed on the CNN and BSIF-based detectors. Despite the specific results vary across these systems, a common thread is the substantial decrease in APCER.  As a matter of fact, integrating the  CC module enables these methods to achieve commendable  results, comparable to state-of-the-art PAD systems (Table \ref{table:sota_err}), depending on the validation set used.

This distinctive characteristic of the CC stems from the inherent nature of presentation attacks. As previously outlined, PA samples lack user-specific characteristics, which are instead present in \textit{bona fide} presentations. Therefore, we can posit that the CC module leverages not only the concept of hierarchical proximity but also its absence.

However, this trend encounters an exception with the MEGVII PAD system. With the addition of the CC module, MEGVII's performance experiences a fine-tuning rather than a drastic boost, subtly enhancing both APCER and BPCER metrics without inducing a severe skew. 
This can be attributed to the already proficient capabilities of MEGVII in distinguishing between \textit{bona fide} and PA samples. This hypothesis is further validated by examining the distribution of the samples; Figure \ref{fig:tsne1} presents the 2D t-SNE visualization of MEGVII and CNN feature embeddings by randomly selecting ten users from the GreenBit 2017 dataset and highlighting the user-specific relationships among the extracted samples. The graphical representation clearly illustrates how MEGVII maintains well-separated distributions for \textit{bona fide} and PA samples, with a solid user-specific clusterization observed for BF feature sets. This characteristic is less pronounced in the case of CNN, where the distributions do not exhibit a similar level of separation (due to the model's simplicity).
Notably, this analysis reveals no direct correlation between the size of the validation set and the presence of common PAIs across training, validation, and testing sets. This suggests that the efficacy of the Closeness Binary Code is not inherently dependent on these factors, indicating its robustness across various experimental conditions. 
A second critical benefit of the CC module lies in its capability to reduce the BPCER while sustaining optimal system performance. This benefit becomes particularly evident when confronting the performances of baseline PAD systems with those combined with the CC module. 
To validate this claim, we conducted further analysis to recalibrate the acceptance threshold of the baseline systems to match the BPCER level of the CC-enhanced systems. 
The outcomes of this experiment, found in Tables \ref{table:BSIFrecalib}, \ref{table:CNNrecalib}, \ref{table:PADUnk21recalib} and \ref{table:MEGVIIrecalib}, show an evident degradation in efficacy against presentation attacks when we adjust the PAD threshold.
 This fact becomes evident when we recognize that the PAD threshold often needs to be significantly lowered from its standard value of 0.5 to match the CC-enhanced system's performance.
Hence, the central role of the CC module becomes apparent: to ensure a low BPCER rate without substantially lowering the security threshold, a compromise that could threaten the system's ability to detect presentation attacks. 

This also addresses the pivotal issue we discussed in our previous research \cite{micheletto2021fingerprint}, where the increase of False Non-Match Rate induced by the PAD emerged as a major factor disrupting performance consistency when combined with verification systems.
Explicitly conceived for integrated systems where templates of the user population are available, the CC module becomes an indispensable asset for maintaining robust, reliable, and secure performance in real-world biometric system deployments. One practical example of such deployment could be within smartphone-based biometric systems. In this context, the CC module could leverage the data captured during the enrollment process, integrating these references to potentially enhance PAD capabilities during subsequent verifications. In fact, modern smartphones, with their computational resources and storage capacity, would be capable of supporting this integration, ensuring that the module operates efficiently with minimal impact on performance. 
For instance, storing a MEGVII template requires only 128 bytes, which is well within the storage capabilities of current devices and justifies the feasibility of maintaining a validation set directly on the smartphone. Additionally, these templates can be securely encrypted using template protection techniques, such as biometric cryptosystems or cancellable biometrics \cite{sarkar2020review}, mitigating privacy concerns and ensuring that sensitive biometric information remains protected. To comprehensively evaluate its potential in this integration scenario, the next section delves into an analysis of the CC module’s effectiveness within biometric systems integrated with PAD capabilities




\begin{table}
\centering
\caption{Comparative analysis of the BSIF performance modifying the acceptance threshold to bring the BPCER of the model to the same value as the CC-enhanced model.}
\label{table:BSIFrecalib}
\resizebox{0.85\linewidth}{!}{
\begin{tabular}{|c|c|c|c|c|} 
\hline
\textbf{\begin{tabular}[c]{@{}c@{}}Validation\\set\end{tabular}} & \textbf{Test set} & \textbf{APCER}  &\begin{tabular}[c]{@{}c@{}}\textbf{APCER@ }\\\textbf{[BPCER=BPCER\textsuperscript{CC}]}\end{tabular} & \textbf{\begin{tabular}[c]{@{}c@{}}new\\THR\end{tabular}} \\ 
\hline
\multirow{3}{*}{\textbf{test21}} & \textbf{test17 [-]} &20.54& 35.39 & 0.07 \\ 
\cline{2-5}
 & \textbf{test19 [M1]} &8.17& 8.01 & 0.51 \\ 
\cline{2-5}
 & \textbf{train19 [BD]} &9.42& 9.75 & 0.45 \\ 
\hline
\multirow{4}{*}{\textbf{test19}} & \textbf{test17 [LE]} &20.54& 41.67 & 0.03 \\ 
\cline{2-5}
 & \textbf{test21 [M1]} &19.27& 62.8 & 0.01 \\ 
\cline{2-5}
 & \textbf{test21ss [M1]} &53.29& 83.86 & 0.01 \\ 
\cline{2-5}
 & \textbf{train19 [-]} &9.42& 15.92 & 0.20 \\ 
\hline
\multirow{4}{*}{\textbf{test17}} & \textbf{test19 [LE]} &8.17& 7.43 & 0.55 \\ 
\cline{2-5}
 & \textbf{test21 [-]} &19.27& 43.94 & 0.08 \\ 
\cline{2-5}
 & \textbf{test21ss [-]} &53.29& 72.24 & 0.08 \\ 
\cline{2-5}
 & \textbf{train19 [-]} &9.42&18.00& 0.16 \\ 
\hline
\multirow{4}{*}{\textbf{train19}} & \textbf{test17 [-]} &20.54& 30.59 & 0.14 \\ 
\cline{2-5}
 & \textbf{test19 [-]} &8.17& 5.31 & 0.70 \\ 
\cline{2-5}
 & \textbf{test21 [BD]} &19.27& 35.04 & 0.17 \\ 
\cline{2-5}
 & \textbf{test21ss [BD]} &53.29& 66.42 & 0.17 \\
\hline
\end{tabular}}
\end{table}

\begin{table}
\centering
\caption{Comparative analysis of the CNN performance modifying the acceptance threshold to bring the BPCER of the model to the same value as the  CC-enhanced model.}
\label{table:CNNrecalib}
\resizebox{0.85\linewidth}{!}{
\begin{tabular}{|c|c|c|c|c|} 
\hline
\textbf{\begin{tabular}[c]{@{}c@{}}Validation\\set\end{tabular}} & \textbf{Test set} & \textbf{APCER}  &\begin{tabular}[c]{@{}c@{}}\textbf{APCER@ }\\\textbf{[BPCER=BPCER\textsuperscript{CC}]}\end{tabular} & \textbf{\begin{tabular}[c]{@{}c@{}}new\\THR\end{tabular}} \\ 
\hline
\multirow{3}{*}{\textbf{test21}} & \textbf{test17 [-]} &24.46&37.25&0.01 \\
\cline{2-5}
 & \textbf{test19 [M1]} &7.92&12.25&0.16\\
\cline{2-5}
 & \textbf{train19 [BD]} &24.33&26.50&0.22\\
\hline
\multirow{4}{*}{\textbf{test19}} & \textbf{test17 [LE]} &24.46&37.25&0.01\\
\cline{2-5}
 & \textbf{test21 [M1]} &13.41&29.67&0.02\\
\cline{2-5}
 & \textbf{test21ss [M1]} &82.76&93.05&0.02\\
\cline{2-5}
 & \textbf{train19 [-]} &24.33&30.75&0.02\\
\hline
\multirow{4}{*}{\textbf{test17}} & \textbf{test19 [LE]} &7.92&9.80&0.34\\
\cline{2-5}
 & \textbf{test21 [-]} &13.41&19.39&0.18\\
\cline{2-5}
 & \textbf{test21ss [-]} &82.76&88.54&0.18\\
\cline{2-5}
 & \textbf{train19 [-]} &24.33&25.17&0.40\\
\hline
\multirow{4}{*}{\textbf{train19}} & \textbf{test17 [-]} &24.46&37.25&0.01\\
\cline{2-5}
& \textbf{test19 [-]} &7.92&15.60&0.05\\
\cline{2-5}
 & \textbf{test21 [BD]} &13.41&26.46&0.04\\
\cline{2-5}
 & \textbf{test21ss [BD]} &82.76&91.79&0.04\\

\hline
\end{tabular}}
\end{table}

\begin{table}
\centering
\caption{Comparative analysis of the PADUnk21 performance modifying the acceptance threshold to bring the BPCER of the model to the same value as the  CC-enhanced model.}
\label{table:PADUnk21recalib}
\resizebox{0.85\linewidth}{!}{
\begin{tabular}{|c|c|c|c|c|} 
\hline
\textbf{\begin{tabular}[c]{@{}c@{}}Validation\\set\end{tabular}} & \textbf{Test set} & \textbf{APCER}  &\begin{tabular}[c]{@{}c@{}}\textbf{APCER@ }\\\textbf{[BPCER=BPCER\textsuperscript{CC}]}\end{tabular} & \textbf{\begin{tabular}[c]{@{}c@{}}new\\THR\end{tabular}} \\ 
\hline
\multirow{3}{*}{\textbf{test21}} & \textbf{test17 [-]} &28.92&28.92& 0.49 \\ 
\cline{2-5}
 & \textbf{test19 [M1]} &18.46&5.64& 0.61 \\ 
\cline{2-5}
 & \textbf{train19 [BD]} &30.17&10.33& 0.60 \\ 
\hline
\multirow{4}{*}{\textbf{test19}} & \textbf{test17 [LE]} &28.92&23.09& 0.53 \\ 
\cline{2-5}
 & \textbf{test21 [M1]} &37.20&26.06& 0.55 \\ 
\cline{2-5}
 & \textbf{test21ss [M1]} &18.41&10.61& 0.55 \\ 
\cline{2-5}
 & \textbf{train19 [-]} &30.17&20.92& 0.54 \\ 
\hline
\multirow{4}{*}{\textbf{test17}} & \textbf{test19 [LE]} &18.46&9.56& 0.56 \\ 
\cline{2-5}
 & \textbf{test21 [-]} &37.20&41.14& 0.47 \\ 
\cline{2-5}
 & \textbf{test21ss [-]} &18.41&21.14& 0.47 \\ 
\cline{2-5}
 & \textbf{train19 [-]} &30.17&43.92& 0.42 \\ 
\hline
\multirow{4}{*}{\textbf{train19}} & \textbf{test17 [-]} &28.92&21.08& 0.54 \\ 
\cline{2-5}
& \textbf{test19 [-]} &18.46&7.76& 0.58 \\ 
\cline{2-5}
 & \textbf{test21 [BD]} &37.20&29.31& 0.53 \\ 
\cline{2-5}
 & \textbf{test21ss [BD]} &18.41&12.76& 0.53 \\

\hline
\end{tabular}}
\end{table}

\begin{table}
\centering
\caption{Comparative analysis of the MEGVII performance modifying the acceptance threshold to bring the BPCER of the model to the same value as the  CC-enhanced model.}
\label{table:MEGVIIrecalib}
\resizebox{0.85\linewidth}{!}{
\begin{tabular}{|c|c|c|c|c|} 
\hline
\textbf{\begin{tabular}[c]{@{}c@{}}Validation\\set\end{tabular}} & \textbf{Test set} & \textbf{APCER}  &\begin{tabular}[c]{@{}c@{}}\textbf{APCER@ }\\\textbf{[BPCER=BPCER\textsuperscript{CC}]}\end{tabular} & \textbf{\begin{tabular}[c]{@{}c@{}}new\\THR\end{tabular}} \\ 
\hline
\multirow{3}{*}{\textbf{test21}} & \textbf{test17 [-]} &0.88& 17.06 & 0.06 \\ 
\cline{2-5}
 & \textbf{test19 [M1]} &0.49& 8.66 & 0.05 \\ 
\cline{2-5}
 & \textbf{train19 [BD]} &0.00&0.00& 0.56 \\ 
\hline
\multirow{4}{*}{\textbf{test19}} & \textbf{test17 [LE]} &0.88& 12.75 & 0.10 \\ 
\cline{2-5}
 & \textbf{test21 [M1]} &2.72& 2.72 & 0.49 \\ 
\cline{2-5}
 & \textbf{test21ss [M1]} &13.62& 13.62 & 0.49 \\ 
\cline{2-5}
 & \textbf{train19 [-]} &0.00&0.00& 0.47 \\ 
\hline
\multirow{4}{*}{\textbf{test17}} & \textbf{test19 [LE]} &0.49& 14.05 & 0.01 \\ 
\cline{2-5}
 & \textbf{test21 [-]} &2.72& 1.42 & 0.62 \\ 
\cline{2-5}
 & \textbf{test21ss [-]} &13.62& 10.16 & 0.62 \\ 
\cline{2-5}
 & \textbf{train19 [-]} &0.00 &0.00& 0.57 \\ 
\hline
\multirow{4}{*}{\textbf{train19}} & \textbf{test17 [-]} &0.88& 8.77 & 0.14 \\ 
\cline{2-5}
& \textbf{test19 [-]} &0.49& 17.89 & 0.00 \\ 
\cline{2-5}
 & \textbf{test21 [BD]} &2.72& 2.72 & 0.49 \\ 
\cline{2-5}
 & \textbf{test21ss [BD]} &13.62& 13.62 & 0.49 \\
\hline
\end{tabular}}
\end{table}

\subsection{Impact of the CC module on integrated results}
Until this section, the Closeness Binary Code analysis and evaluation have predominantly focused on assessing its performance within independent PAD systems. However, this approach does not provide a comprehensive or realistic understanding of CC's operational dynamics. It overlooks the critical fact that PAD systems are not standalone entities: they are designed to work within the context of broader biometric recognition systems, which they aim to safeguard.
However, as previous studies have highlighted \cite{chingovska2019evaluation, micheletto2021fingerprint}, the integration of PAD systems can significantly impact the overall performance of the biometric recognition system. The most noticeable issue is the False Non-Match Rate (FNMR) increase in the integrated systems. Such rise is influenced by various factors, including the efficiency of the PAD, its selected operational point, and the expected probability of encountering presentation attacks.
Hence, the goal of this section is to assess the CC's performance within the recognition system context to provide insights into its effectiveness in enhancing overall system security, robustness, and reliability.

For this purpose, we exploited Bio-WISE\footnote{https://livdet.pythonanywhere.com/}, a simulation tool designed for the integration of biometric recognition systems \cite{micheletto2021fingerprint}.
Bio-WISE works by taking as inputs the individual Receiver Operating Characteristic (ROC) curves of the comparator and of the PAD. Then it provides the ROC curve of the fused system, considering two fundamental parameters: the prior probability of being subjected to a presentation attack, denoted by '$w$', and the specific operating point chosen for the PAD. For consistency with the previous analysis, we have chosen to maintain the same PAD operating point ($thr=0.5$). As for the probability of an attack, we examined two key scenarios:
\begin{itemize}
    \item Absence of a Presentation Attack ($w=0$): This analysis aims to demonstrate the improvement of the False Non-Match Rate on the integrated system when the Closeness Binary Code is employed. We acknowledge that using a PAD when $w=0$ might not be typically practical. However, this scenario offers valuable information on the potential benefits of incorporating the CC module even in low-risk settings.
    \item A moderately non-zero attack probability ($w=0.1$): this scenario assess the impact of PAs at a realistic level, emphasizing the importance of PAD in enhancing system security. The integration of the CC module showcases further improvements in detection capabilities, highlighting the utility of the proposed add-on as an integral part of the PAD within the broader recognition system.
\end{itemize}

\begin{figure*}[!h]

   \centering
      \subfigure[]{\includegraphics[width=.23\textwidth]{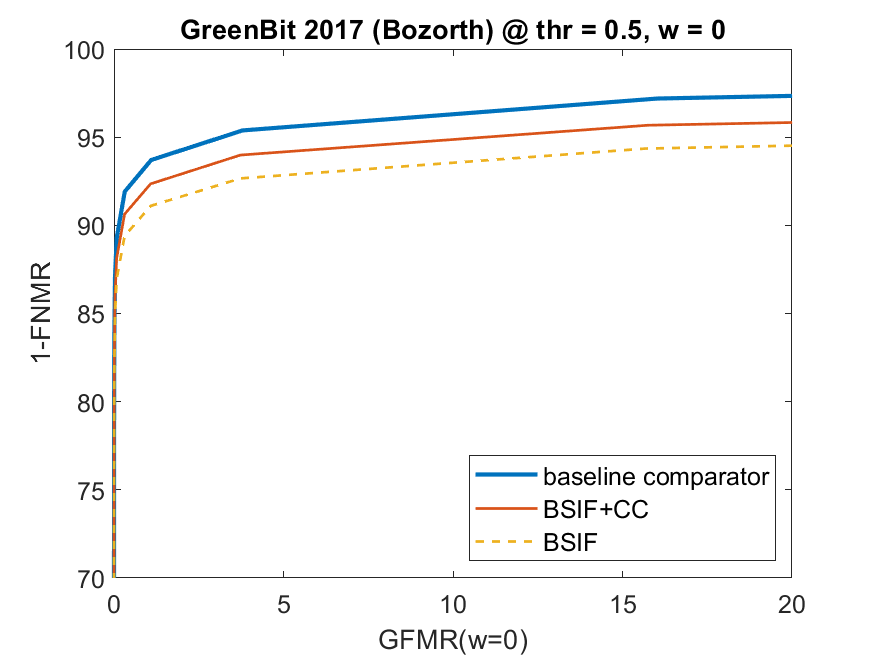}}
     \subfigure[]{\includegraphics[width=.23\textwidth]{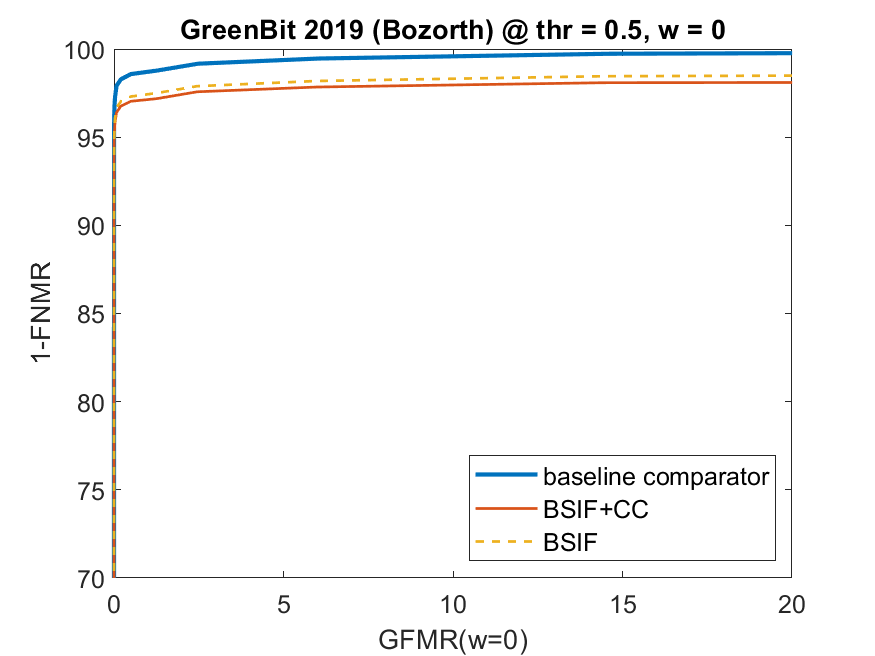}}
   \subfigure[]{\includegraphics[width=.23\textwidth]{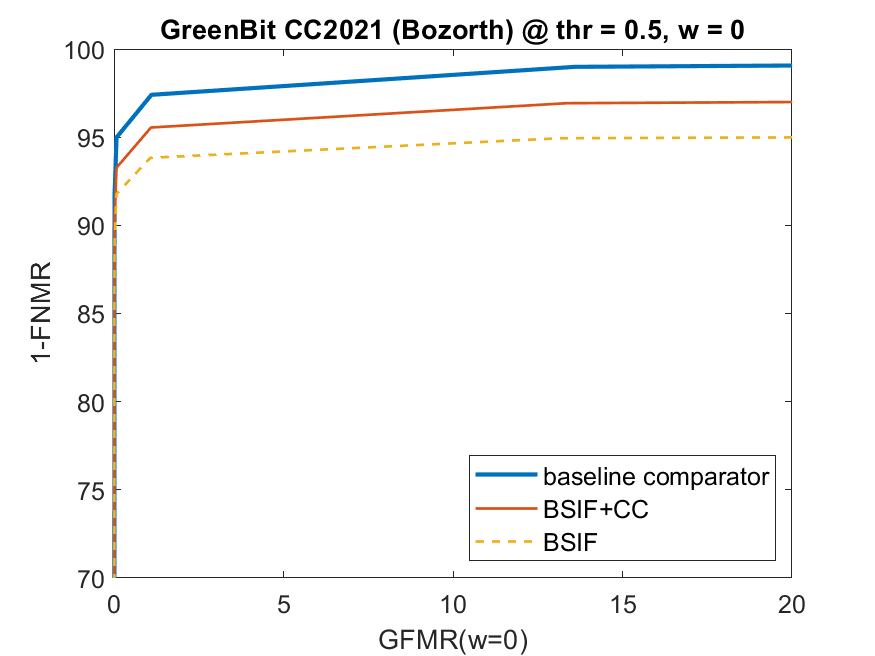}}
     \subfigure[]{\includegraphics[width=.23\textwidth]{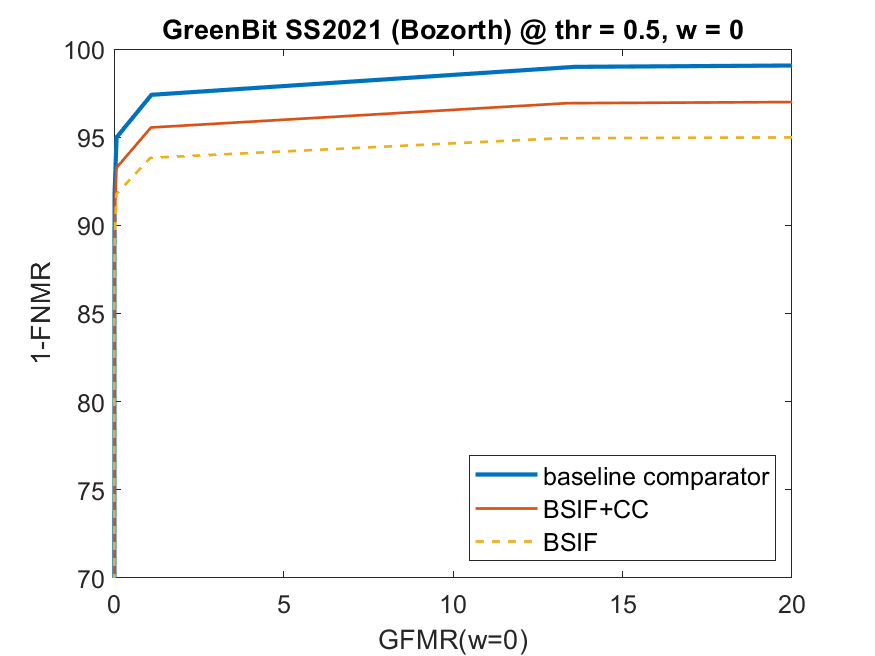}}
           \subfigure[]{\includegraphics[width=.23\textwidth]{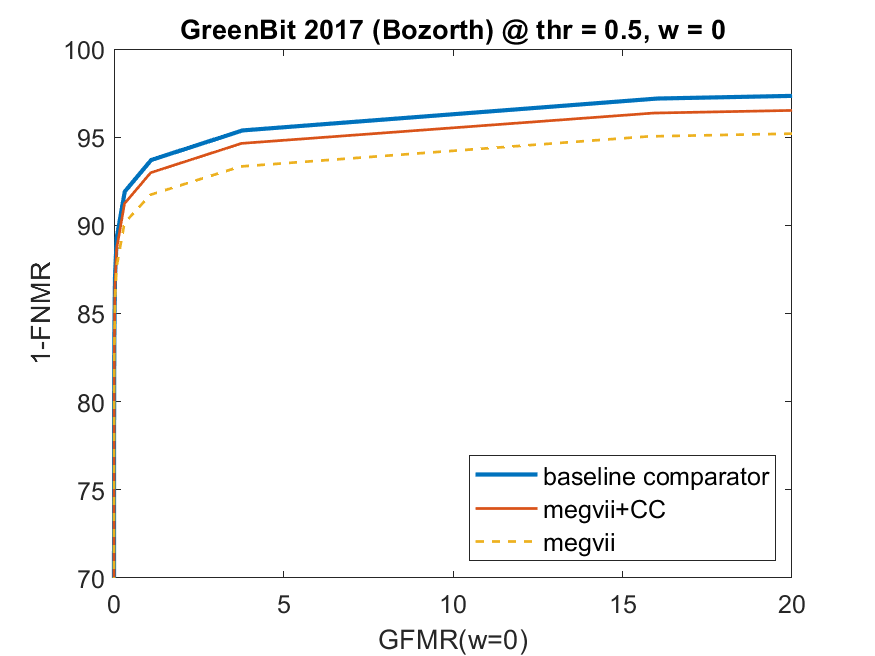}}
     \subfigure[]{\includegraphics[width=.23\textwidth]{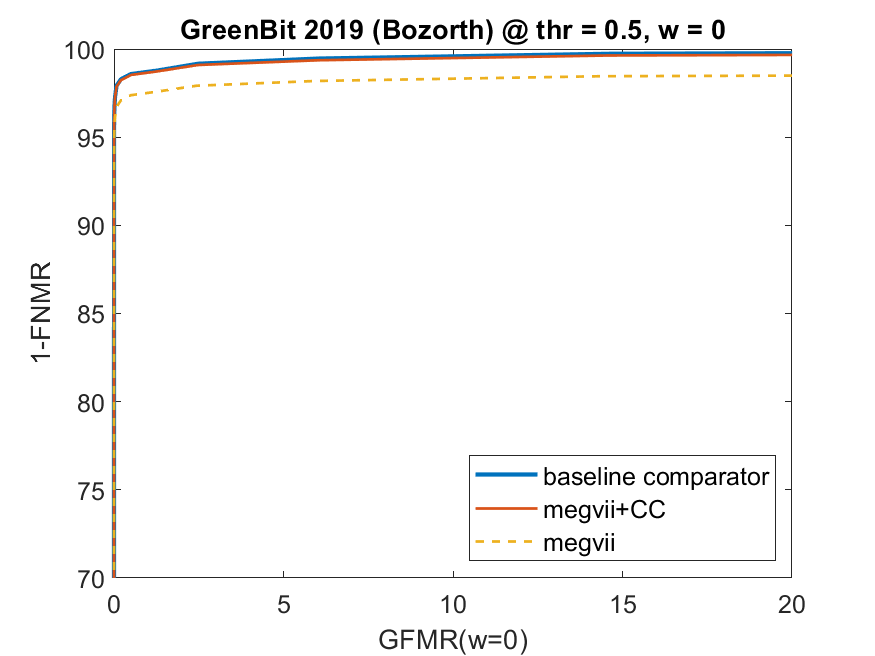}}
   \subfigure[]{\includegraphics[width=.23\textwidth]{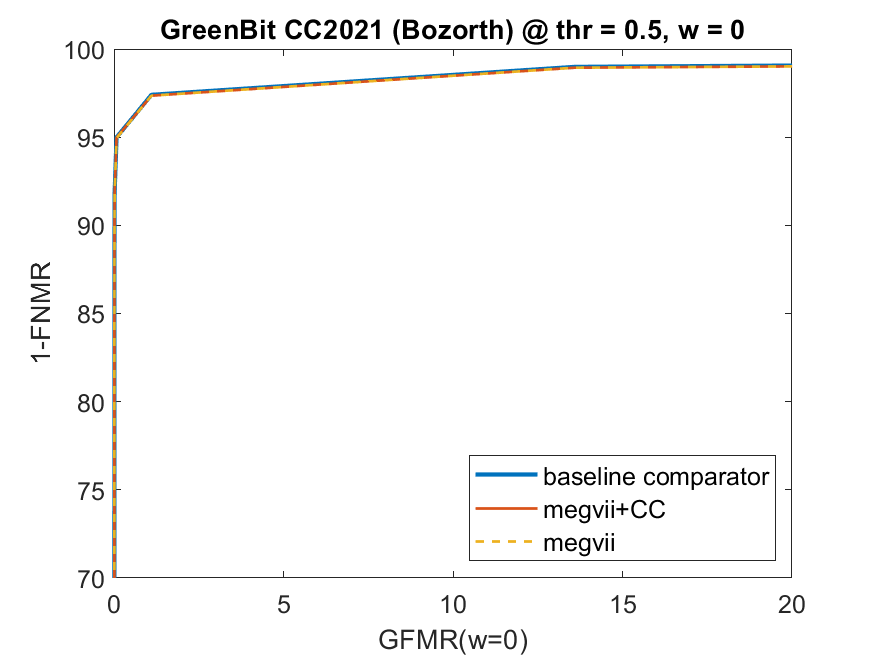}}
     \subfigure[]{\includegraphics[width=.23\textwidth]{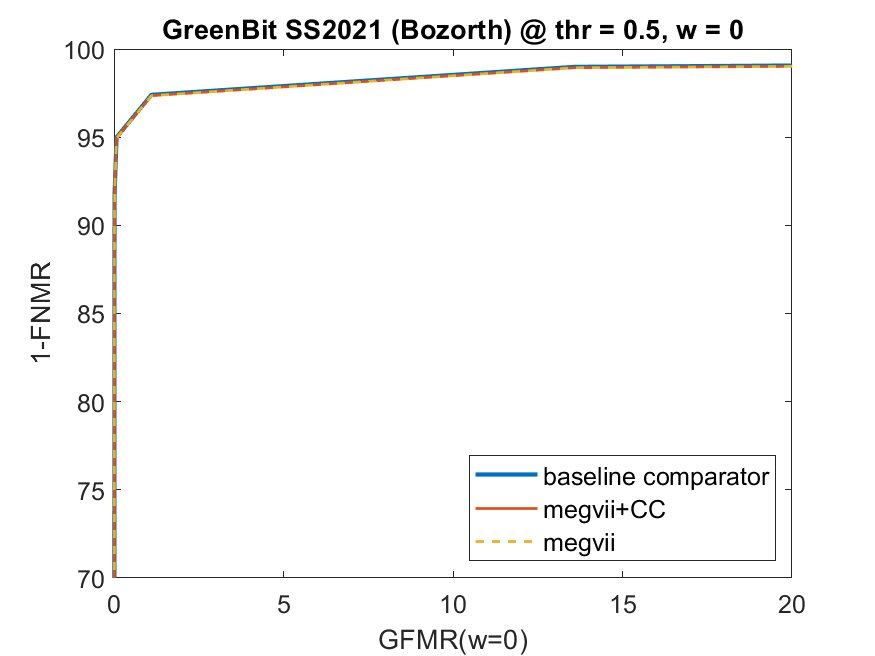}}
\caption{Comparison of ROCs between a baseline comparator using Bozorth3 and its integration with a BSIF (a-b-c-d) or MEGVII (e-f-g-h)  Presentation Attack Detection (PAD) system on the entire test set, where the presentation attack probability ($w$) is set to 0.}
\label{fig:integ0}
\vspace{-10pt}
\end{figure*}

\begin{figure*}[!h]

   \centering
      \subfigure[]{\includegraphics[width=.23\textwidth]{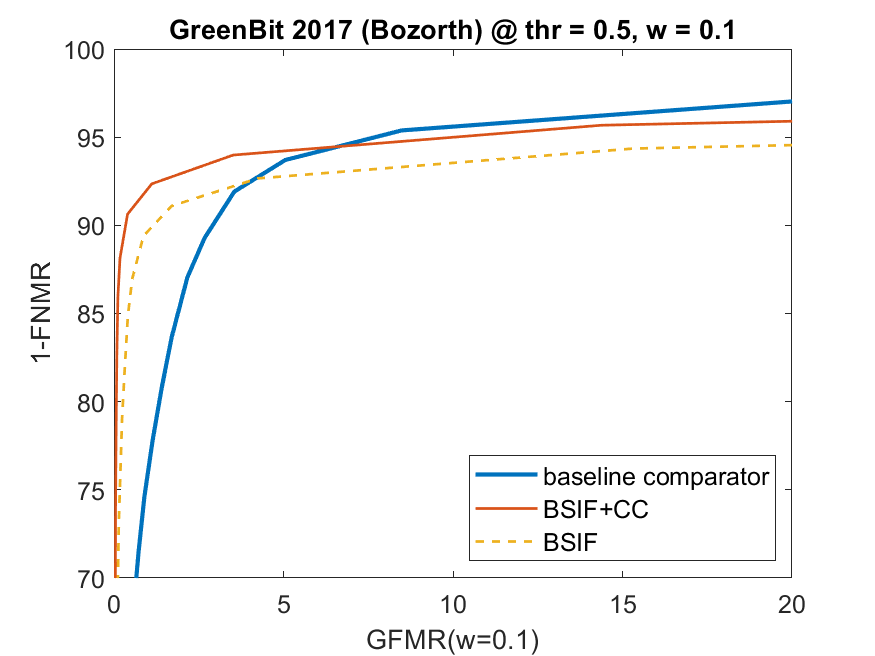}}
     \subfigure[]{\includegraphics[width=.23\textwidth]{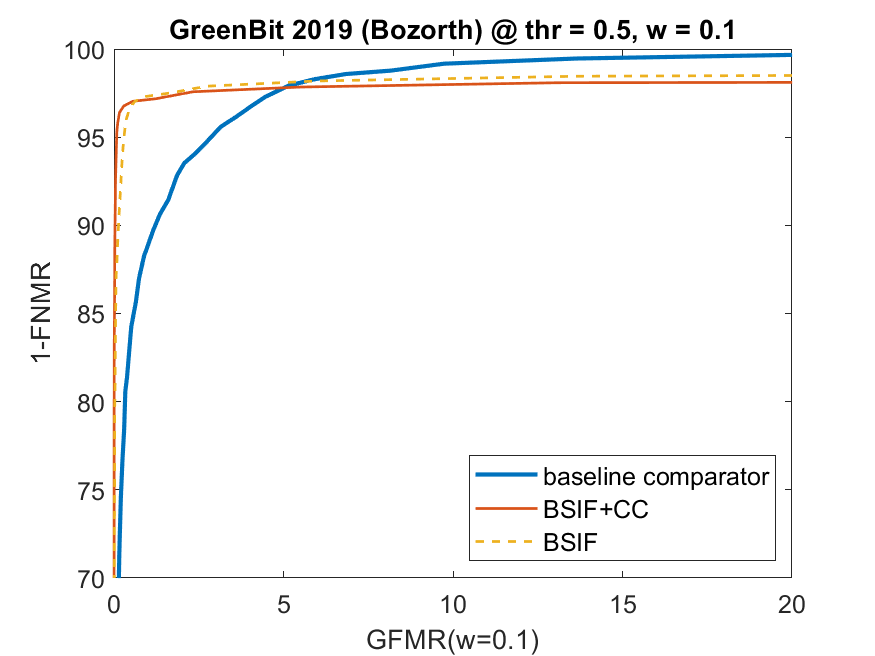}}
   \subfigure[]{\includegraphics[width=.23\textwidth]{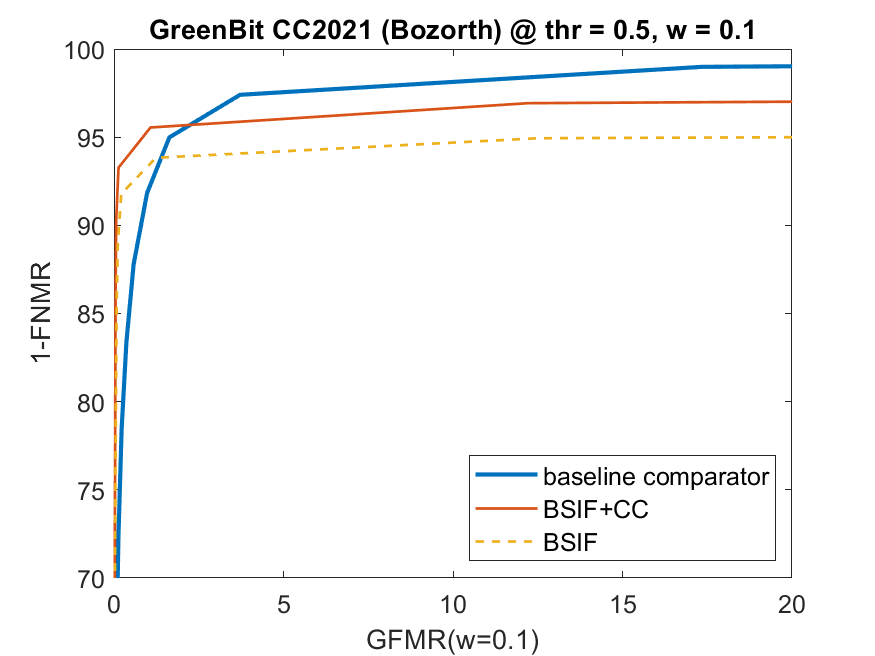}}
     \subfigure[]{\includegraphics[width=.23\textwidth]{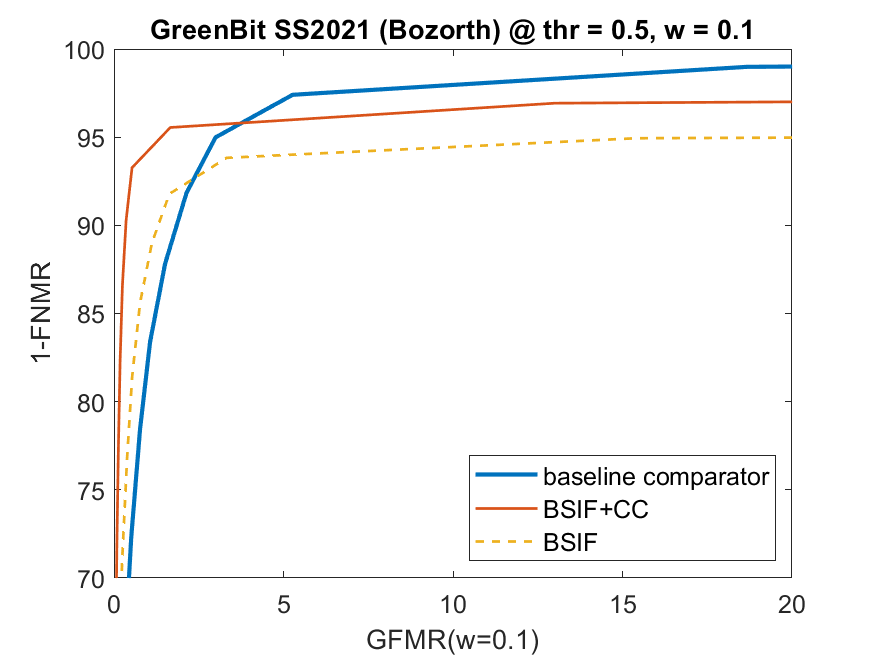}}
           \subfigure[]{\includegraphics[width=.23\textwidth]{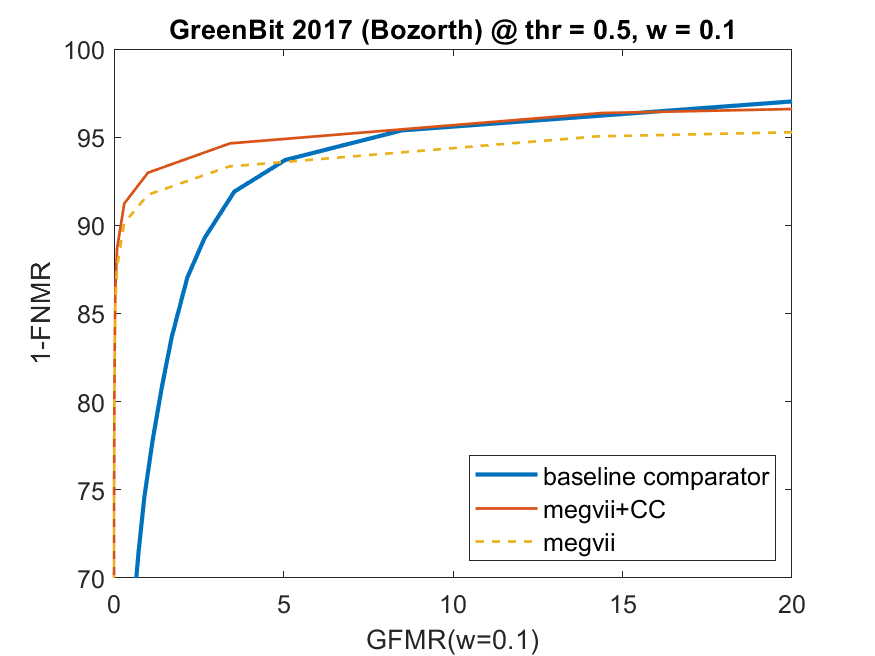}}
     \subfigure[]{\includegraphics[width=.23\textwidth]{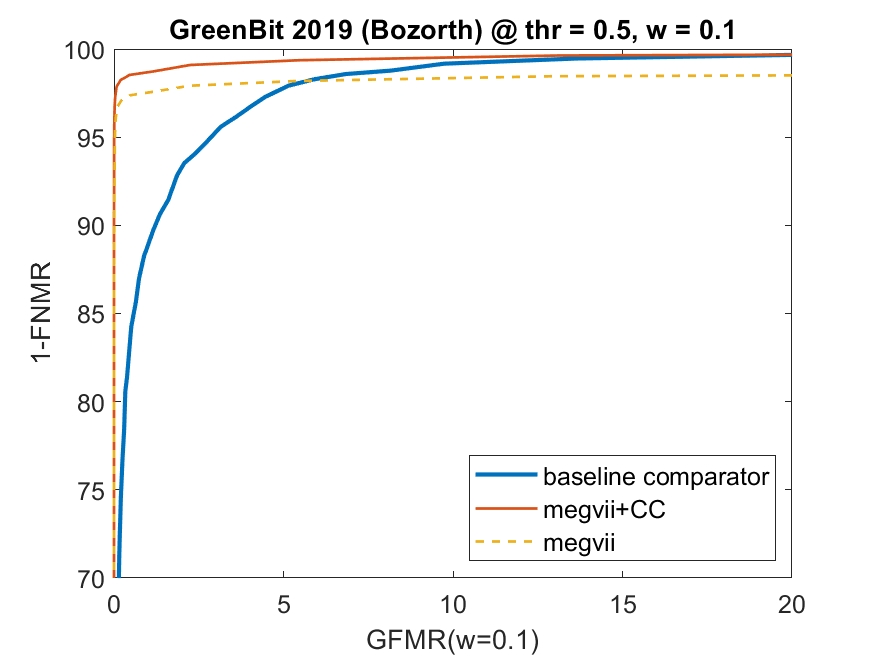}}
   \subfigure[]{\includegraphics[width=.23\textwidth]{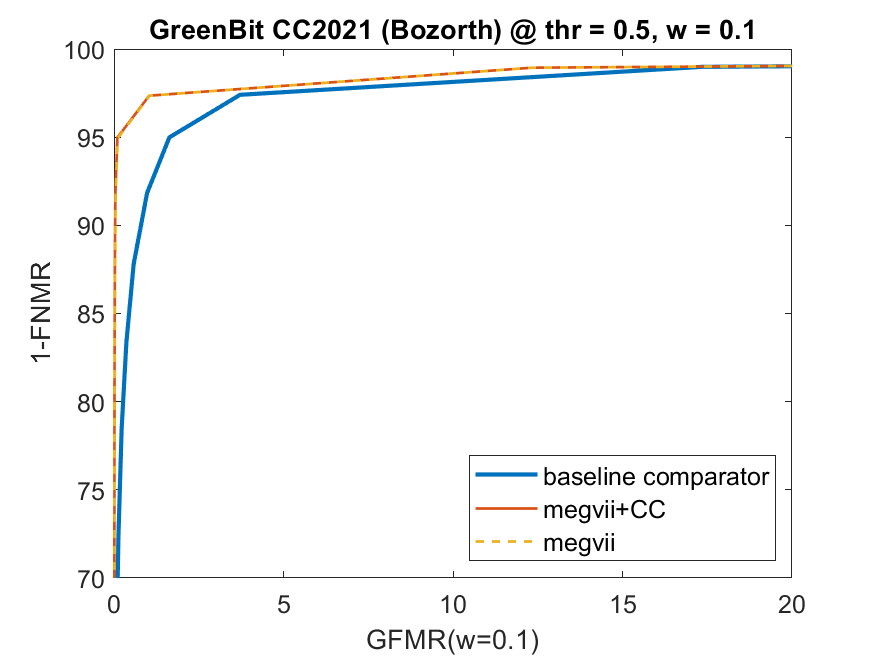}}
     \subfigure[]{\includegraphics[width=.23\textwidth]{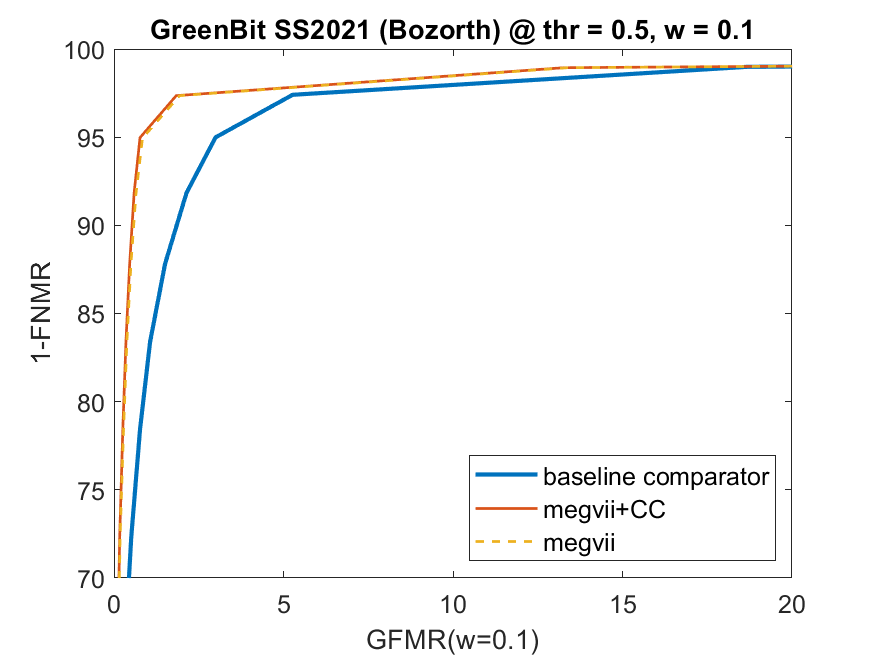}}
\caption{Comparison of ROCs between a baseline comparator using Bozorth3 and its integration with a BSIF (a-b-c-d) or MEGVII (e-f-g-h) Presentation Attack Detection (PAD) system on the entire test set, where the presentation attack probability ($w$) is set to 0.1.}
\label{fig:integ03}
\vspace{-10pt}
\end{figure*}

For the sake of space, we selected two prominent PAD systems that exemplify the main types examined in the paper: BSIF, representing a handcrafted approach, and MEGVII, representing a deep learning approach. We embedded them with the standard NIST Bozorth3\footnote{https://www.nist.gov/services-resources/software/nist-biometric-image-software-nbis} comparator. 
The results are shown in Figures \ref{fig:integ0} and \ref{fig:integ03}, where we illustrate the ROC curves for the integrated system with and without the Closeness Binary Code, which was customized using the LivDet 2019 training set as validation set.
It is important to note that the x-axis of the graph represents the GFMR (Global False Match Rate), as defined in Ref. \cite{micheletto2021fingerprint}, p. 5343, Eq. 23. The GFMR is a weighted sum of the False Match Rate (FMR) and the Impostor Attack Presentation Acceptance Rate (IAPAR), where the probability of attack $w$ determines the weighting. This measure is essential in evaluating the integrated system performance since it considers both non-mated trials and misclassified PAs.

The results allow us to appreciate the benefit of the CC module on the integrated AFIS in both scenarios examined. In fact, for any operating point of the comparator, the PAD with CC maintains lower error rates than the PAD without CC.

In particular, when $w=0$ (Fig. \ref{fig:integ0}), the CC-enhanced PAD consistently performs better than the PAD without the Closeness Binary Code. However, by the inherent nature of a PAD, it introduces an additional error on the FNMR, making its performance fall short of the baseline comparator. This finding highlights the underlying trade-off in biometric security: while the PAD significantly increases the system's resilience against presentation attacks, it does so at the cost of an increase in FNMR. Nevertheless, introducing the Closeness Binary Code into the PAD mitigates this increase, enhancing the integrated system's overall performance in most cases.

In the second scenario (Fig. \ref{fig:integ03}), as the risk of attacks cannot be neglected ($w=0.1$), the comparator, lacking the necessary defences against such threats, exhibits poorer performance, as evidenced by its higher GFMR. This illustrates the importance of a PAD, especially in settings where presentation attacks are more likely. 
Even in this case, the integrated system equipped with the CC module offers a more secure and reliable solution.


\subsection{Comparison with SOTA domain generalization methods}
The goal of this section is to evaluate the Closeness Code module by comparing its performance with other SOTA techniques that, similar to our approach, aim to enhance the generalization capabilities of PAD systems through additional strategies. Some methods, such as Channel Feature Denoising PAD (CFD-PAD) \cite{liu2022fingerprint}, implement an add-on by using channel denoising to refine feature extraction, while others, such as the Adversarial Liveness Detector (ALD)\cite{galli2023} based on Adversarial Data Augumentation (ADA) or the Universal Material Generator (UMG) \cite{chugh2020fingerprint}, leverage additional training data (adversarial perturbations for ALD and style transfer for UMG) to improve generalization without employing separate add-on components.

Table \ref{table:sota_acer} presents the ACER values for each PAD system, including our methods with and without the CC module, and SOTA baseline systems combined with their respective enhancements. We also show the percentage change ($\Delta$ ACER) to provide a normalized and fair comparison across different approaches. All evaluations were conducted on the LivDet 2017 dataset, using LivDet 2019 as the validation set for the LUT table.

The results demonstrate that PAD systems integrated with the CC module achieve significant percentage improvements in ACER, often showing comparable or superior performance to other advanced methods. The strength of the CC module lies in its efficiency, as it introduces minimal additional computational demands while enhancing performance. In fact, when evaluating methods aimed at improving generalization, it is essential to balance computational complexity, processing time, and resource requirements against the performance gains achieved. The CC module effectively strikes this balance, delivering improvements consistent with those of more computationally intensive SOTA techniques. The detailed analysis of the computational impact of the CC module is presented in the upcoming Section \ref{sec:time_anal}. Finally, it is worth remarking that the comparison presented here serves merely as a benchmark example; the CC module’s adaptable design means it could be applied to any state-of-the-art PAD system, potentially enhancing performance and generalization capabilities across a wide range of configurations.

\begin{table*}[]
\centering
\caption{Comparison of the ACER decrease on test2017 obtainable by the closeness binary code module with respect to other state-of-the-art techniques proposed to increase the generalization capabilities in the PAD task. The techniques compared are Adversarial Data augmentation (ADA) \cite{galli2023}, Universal Material Generator (UMG) \cite{chugh2020fingerprint} and Channel-wise Feature Denoising (CFD) \cite{liu2022fingerprint}.}
\label{table:sota_acer}
\resizebox{\linewidth}{!}{
\begin{tabular}{|c|cccccccccc|cccccc|}
\hline
\multirow{2}{*}{}   & \multicolumn{10}{c|}{\textbf{Closeness Binary Code add-on {[}proposed{]}}}& \multicolumn{6}{c|}{\textbf{State-of-the-art generalization methods}}\\ \cline{2-17} 
                    & \multicolumn{1}{c|}{\textbf{PADunk}} & \multicolumn{1}{c|}{\textbf{\begin{tabular}[c]{@{}c@{}}PADunk\\ +\\ CC\end{tabular}}} & \multicolumn{1}{c|}{\textbf{Megvii}} & \multicolumn{1}{c|}{\textbf{\begin{tabular}[c]{@{}c@{}}Megvii\\ +\\ CC\end{tabular}}} & \multicolumn{1}{c|}{\textbf{CNN}} & \multicolumn{1}{c|}{\textbf{\begin{tabular}[c]{@{}c@{}}CNN\\ +\\ CC\end{tabular}}} & \multicolumn{1}{c|}{\textbf{BSIF}} & \multicolumn{1}{c|}{\textbf{\begin{tabular}[c]{@{}c@{}}BSIF\\ +\\ CC\end{tabular}}} & \multicolumn{1}{c|}{\textbf{LBP}} & \textbf{\begin{tabular}[c]{@{}c@{}}LBP\\ +\\ CC\end{tabular}} & \multicolumn{1}{c|}{\textbf{EfficientNet}} & \multicolumn{1}{c|}{\textbf{\begin{tabular}[c]{@{}c@{}}EfficientNet\\ +\\ ADA\end{tabular}}} & \multicolumn{1}{c|}{\textbf{FSB}} & \multicolumn{1}{c|}{\textbf{\begin{tabular}[c]{@{}c@{}}FSB\\ +\\ UMG\end{tabular}}} & \multicolumn{1}{c|}{\textbf{MobileNet}} & \textbf{\begin{tabular}[c]{@{}c@{}}MobileNet\\ +\\ CFD\end{tabular}} \\ \hline
\textbf{ACER}       & \multicolumn{1}{c|}{16.84}           & \multicolumn{1}{c|}{3.11}                                                             & \multicolumn{1}{c|}{1.56}            & \multicolumn{1}{c|}{0.855}                                                            & \multicolumn{1}{c|}{19.29}        & \multicolumn{1}{c|}{10.94}                                                        & \multicolumn{1}{c|}{11.71}         & \multicolumn{1}{c|}{9.11}                                                           & \multicolumn{1}{c|}{19.42}        & 5.39                                                          & \multicolumn{1}{c|}{10.81}                 & \multicolumn{1}{c|}{10.51}                                                                   & \multicolumn{1}{c|}{3.32}         & \multicolumn{1}{c|}{2.58}                                                           & \multicolumn{1}{c|}{3.54}               & 2.59                                                                 \\ \hline
\textbf{$\Delta$ ACER} & \multicolumn{2}{c|}{-81.53}                                                                                                  & \multicolumn{2}{c|}{-45.19}                                                                                                  & \multicolumn{2}{c|}{-43.31}                                                                                            & \multicolumn{2}{c|}{-22.20}                                                                                              & \multicolumn{2}{c|}{-72.25}                                                                       & \multicolumn{2}{c|}{-2.78}                                                                                                                & \multicolumn{2}{c|}{-22.29}                                                                                             & \multicolumn{2}{c|}{-26.84}                                                                                    \\ \hline
\end{tabular}}
\end{table*}

\subsection{Time analysis}
\label{sec:time_anal}
In the context of deploying practical biometric systems, computational efficiency is a vital factor. Therefore, we conducted a final analysis to examine the execution times associated with implementing the CC module. We aimed to dispel the notion that adding the CC module might introduce a computationally heavy overhead.

The computation times reported in this analysis, as shown in Table \ref{table:executionTimes}, span the period from the receipt of the sample to be classified up to its final classification. Thus, these times include the calculation of distances with the samples in the dataset, the computation of its Closeness Binary Code sequence, and the time needed to make the final decision.

\begin{table}
\centering
\caption{Average increase in PAD classification times of an input sample due to the CC module.}
\label{table:executionTimes}
\begin{tabular}{|c|c|c|c|c|c|} 
\hline
\textbf{Method} & \textbf{Megvii} & \textbf{BSIF}  & \textbf{CNN} & \textbf{PADUnk21} \\ 
\hline
\textbf{Avg Time (ms)} & 0.2481 & 0.3497  & 0.2558 & 1.5172 \\
\hline
\end{tabular}
\end{table}

Overall, the computational burden introduced by the CC module is minimal. Most methods exhibit an average computation time under 0.4 ms, with PADUnk21 being a notable exception with an average of 1.52 ms. Nevertheless, even in the PADUnk21 case, the added processing time remains acceptable and not a significant deterrent to implementing the CC module in a real-world system.

In summary, these findings reinforce the argument for the utility of the CC module: it significantly enhances the efficacy of PAD systems while maintaining a minimal computational burden, making it an appealing choice for integrated biometric system deployments.

\section{Conclusions}
\label{ref:concl}
The Closeness binary Code module is an add-on to existing PADs based on the concept of hierarchical proximity between \textit{bona fide} embeddings of the same user and exploits the AFIS templates to improve the PAD performance.

The integration of this module to different state-of-the-art AI-based PADs, both white-box and black-box, allowed us to demonstrate the effectiveness of exploiting the user-specific effect to distinguish \textit{bona fide} and presentation attacks.
This performance increase is relevant when the distributions of the PA and BF samples in the feature space significantly overlap: this situation often occurs in case of never-seen-before PAIs, which turns out as a favorable event for the proposed add on. 
Our experiments highlighted the CC add on's ability to exploit user-specific information within the AFIS framework, without requiring user-specific PA data for the CC's lookup table design and keeping computation times limited. 




To sum up, the low computational complexity and the high-performance improvement reported, we believe that the proposed ``closeness binary code'' add-on is a significant feature to enhance AFIS-integrated PADs by taking advantage of enrolled users' information and by avoiding the re-training or fine-tuning of the PAD in any way.


%




\bibliographystyle{IEEEtran}
%

\bibliography{main}

\end{document}